# Robust Local Search for Solving RCPSP/max with Durational Uncertainty

**Na Fu**                                                    na.fu.2007@phdis.smu.edu.sg
**Hoong Chuin Lau**                                                    hclau@smu.edu.sg
**Pradeep Varakantham**                                                    pradeepv@smu.edu.sg
*School of Information Systems,*
*Singapore Management University,*
*80 Stamford Road, 178902 Singapore*

**Fei Xiao**                                                    feixiao@gmail.com
*Google Inc.*
*1600 Amphitheatre Parkway Mountain View,*
*CA 94043 USA*

## Abstract

Scheduling problems in manufacturing, logistics and project management have frequently been modeled using the framework of Resource Constrained Project Scheduling Problems with minimum and maximum time lags (RCPSP/max). Due to the importance of these problems, providing scalable solution schedules for RCPSP/max problems is a topic of extensive research. However, all existing methods for solving RCPSP/max assume that durations of activities are known with certainty, an assumption that does not hold in real world scheduling problems where unexpected external events such as manpower availability, weather changes, etc. lead to delays or advances in completion of activities. Thus, in this paper, our focus is on providing a scalable method for solving RCPSP/max problems with durational uncertainty. To that end, we introduce the robust local search method consisting of three key ideas: (a) Introducing and studying the properties of two decision rule approximations used to compute start times of activities with respect to dynamic realizations of the durational uncertainty; (b) Deriving the expression for *robust makespan* of an execution strategy based on decision rule approximations; and (c) A robust local search mechanism to efficiently compute activity execution strategies that are robust against durational uncertainty. Furthermore, we also provide enhancements to local search that exploit temporal dependencies between activities. Our experimental results illustrate that robust local search is able to provide robust execution strategies efficiently.

## 1. Introduction

Research in scheduling has typically considered problems with deterministic durations. In real-world scheduling problems, unexpected external events such as manpower availability, weather changes, etc. lead to uncertainty about durations of activities. There has been a growing interest to account for such data uncertainty (Herroelen & Leus, 2005; Beck & Wilson, 2007; Rodríguez, Vela, Puente, & Hernández-Arauzo, 2009) while providing optimized schedules. This paper also focuses on this important issue of durational uncertainty in scheduling problems. More specifically, we consider scheduling problems where there are complex resource constraints and temporal dependencies between activities.

There are broadly two approaches for tackling scheduling problems with durational uncertainty. One is to adopt a hybrid of proactive and reactive methods, e.g., the work of Vonder, Demeulemeester, and Herroelen (2007), where an initial baseline schedule is computed offline, which is then modified (if required) during execution reactively based on the occurrence of external events. The second approach, e.g., the paper by Mohring and Stork (2000), is to design schedule policies that provide online decision rules such that at time $t$, the policy decides which task(s) may start and which resource(s) to assign. In this paper, we adopt the latter approach and focus on the computation of a robust schedule policy.

From the computational perspective, stochasticity adds a great deal of complexity to the underlying deterministic scheduling problem. For example, in the infinite-resource project scheduling problem where processing times have two possible discrete values, the problem of computing the expected makespan (or any point on the cumulative distribution of the optimal makespan), is #P-hard (Hagstrom, 1988; Möhring, 2001). It has also been shown that for the scheduling problem $1|stoch\ p_j; d_j = d|E[\sum w_j U_j]$, the problem of computing a policy (i.e., execution strategy) maximizing the probability that some job completes exactly at the deadline is PSPACE-hard (Dean, Goemans, & Vondrák, 2004). Daniels and Carrillo (1997) consider a one-machine scheduling problem with probabilistic durations, with an objective to capture the likelihood that a schedule yields actual performance no worse than a given target level. This has been shown to be NP-hard even though the underlying deterministic problem can be solved in polynomial time.

The concrete problem of interest in this paper is the Resource Constrained Project Scheduling Problem with minimum and maximum time lags (abbrev. RCPSP/max), which is of great importance in manufacturing, logistics and project management. Though these problems have been shown to be NP-Hard (Bartusch, Mohring, & Radermacher, 1988), local search based techniques (Demeulemeester & Herroelen, 2002) have achieved great success in solving these problems. Taking a cue from this and the recent advancements in robust optimization, we propose a robust local search method for solving the RCPSP/max problem under durational uncertainty with a risk management perspective. More precisely, we (a) employ concepts from robust optimization to compute the *robust* makespan with proven success probability (or risk of failure) for an execution strategy; and (b) then use local search methods for computing an execution strategy that seeks to minimize this *robust makespan*.

A recent approach (Beck & Wilson, 2007) provides techniques to compute the robust baseline schedule from a risk management perspective, where durations of activities are modeled as random variables. Given a value $0 < \alpha \leq 1$, they were interested to compute a schedule with minimal (probabilistic) makespan where the probability of successful execution is at least $1 - \alpha$ over all realizations of the durational uncertainty. The main contribution there was to derive a lower bound for the $\alpha$-makespan of a given schedule by solving a deterministic problem. They considered the Job-shop Scheduling Problem (JSP) that represents a special case of RCPSP/max (which is the problem of interest in this paper).

Unlike in JSPs, there are complex resource constraints and activity dependencies in RCPSP/max problems with durational uncertainty. To account for these, we compute an execution strategy (also known commonly as schedule policy) called Partial Order Schedule (POS) instead of a schedule. We combine techniques from robust optimization with classical local search to compute a POS that minimizes the *robust makespan*. The robust makespan is a value for which the probability of realized makespan for any schedule (derived from POS) does not exceed it is greater than $(1 - \varepsilon)$, over all realizations of uncertainty. Thus, we compute an upper bound on makespan values as opposed to lower bound computation in the work of Beck and Wilson (2007).

More specifically, we make three key contributions in this paper. Firstly, we introduce two decision rule approximations to define expressions for start times of activities based on random variables used to represent the durational uncertainties: (a) Segregated Linear Approximation(SLA) and (b) Generalized Non-Linear Approximation (GNLA). Secondly, we derive expressions for the upper bound on robust makespan by employing the one sided Chebyshev's inequality on the decision rule approximations above. Finally, we perform local search for an execution strategy using the robust makespan upper bound. We also provide enhancements that consider feedback about robustness of execution strategies to improve the performance of local search.

In order to demonstrate the effectiveness of our methods, we evaluate the performance on benchmark problem sets of RCPSP/max and Job-shop Scheduling Problems (JSPs) with durational uncertainty. Furthermore, we make an in house comparison amongst various enhancements developed in this paper. Finally, due to the absence of competing algorithms for solving RCPSP/max problems and to provide an indication of the performance provided by robust local search, we compare against the existing best solver for JSPs with durational uncertainty.

In the next section, we present a brief background of the models and solution concepts referred to in this paper. We then present the decision rule approximations in Section 3 and the computation of robust makespan upper bound in Section 4. The detailed description of robust local search and its enhancements are provided in Section 5 and Section 6. Finally, the experimental setup and results are provided in Section 7.

## 2. Preliminaries

In this section, we briefly describe the notations along with the scheduling models and robust optimization concepts of relevance to this paper.

### 2.1 Definitions and Notations

As given by Ben-Tal and Nemirovski (2002), we also classify the variables in a stochastic optimization problem into 2 types: *Adjustable* and *Non-Adjustable variables*.

**Definition 1.** *Non-Adjustable variables are a priori decisions that must be made before the actual realization of the uncertainty.*

**Definition 2.** *Adjustable variables (also known as recourse variables) are 'wait-and-see' variables that can adjust themselves when part of the uncertain data become known.*

For example, in a scheduling problem such as RCPSP with uncertain task durations, the non-adjustable variables will represent the execution policy, e.g., the POS proposed by Policella, Smith, Cesta, and Oddi (2004), that need to be constructed a priori, while the adjustable variables are associated with the actual start times of the tasks, which will be set with respect to the execution policy and dynamic realizations of uncertainty.

A random variable will be denoted by $\tilde{x}$ and bold face lower case letters such as $\mathbf{x}$ represent vectors.

## 2.2 RCPSP/max

We now describe the deterministic RCPSP/max scheduling problem along with the extension to handle durational uncertainty. We also explain the execution policy for an uncertain duration extension of the RCPSP/max.

### 2.2.1 Deterministic RCPSP/max

The RCPSP/max problem (Bartusch et al., 1988) consists of $N$ activities $\{a_1, a_2..., a_N\}$, where each activity $a_j$ ($j = 1, ...N$) is to be executed for a certain amount of time units without preemption. Each activity $a_j$ has a fixed *duration* or *processing time* $d_j$, which is assumed to be a non-negative real number or non-negative integer number. In addition, dummy activities $a_0$ and $a_{N+1}$ with $d_0 = d_{N+1} = 0$ are introduced to represent the beginning and the completion of the project, respectively.

A start time *schedule* $\boldsymbol{ss}$ is an assignment of start times to all activities $a_1, a_2..., a_N$, i.e. a vector $\boldsymbol{ss} = (st(a_1), st(a_2), ...st(a_N))$ where $st(a_i)$ represents the start time of activity $a_i$ and $st(a_0)$ is assumed to be 0. Let $et(a_i)$ be the end time of activity $a_i$. Since durations are deterministic and preemption is not allowed, we then have

$$st(a_i) + d_i = et(a_i). \tag{1}$$

And the project *makespan* which is also the start time of the final dummy activity $st(a_{N+1})$ equals

$$st(a_{N+1}) = max_{i=1,...N} et(a_i). \tag{2}$$

Schedules are subject to two kinds of constraints, *temporal constraints* and *resource constraints*. Temporal constraints restrict the time lags between activities. A *minimum time lag* $T_{ij}^{min}$ between the start time of two different activities $a_i$ and $a_j$ says that

$$st(a_j) - st(a_i) \geq T_{ij}^{min} \tag{3}$$

Specially, $T_{ij}^{min} = 0$ means that activity $a_j$ cannot be started before activity $a_i$ begins. A *maximum time lag* $T_{ij}^{max}$ between the start time of two different activities $a_i$ and $a_j$ says that

$$st(a_j) - st(a_i) \leq T_{ij}^{max} \tag{4}$$

$T_{ij}^{max} = 0$ means that activity $a_j$ cannot be started after activity $a_i$ begins.

In this definition, time lags connect start times of two related activities, known as *start-to-start* time lags. *start-to-end, end-to-end, end-to-start* time lags can be easily transformed to the general *start-to-start* time lags for the deterministic case as given by Bartusch et al. (1988). A schedule $\boldsymbol{ss} = (st(a_1), st(a_2), ...st(a_N))$ is *time feasible*, if all the time lag constraints are satisfied at the start times $st(a_i)$ ($i = 1, ...N$).

A resource unit is reusable and available for another activity once it is no longer used by the current activity. Each type of resource has a limited capacity, $C_k$ ($k = 1, 2..., K$) units. Each activity $a_i$ requires $r_{ik}$ units of resource of type $k$ where $k = 1, 2..., K$. Let $A(t) = \{i \in \{1, 2...N\} | st(a_i) \leq t \leq et(a_i)\}$ be the set of activities which are being processed at time instant $t$. A schedule is *resource feasible* if at each time instant $t$, the total demand for a resource $k$ does not exceed its capacity $C_k$, i.e.

$$\sum_{i \in A(t)} r_{ik} \leq C_k. \tag{5}$$

A schedule $\boldsymbol{ss}$ is called *feasible* if it is both time and resource feasible. The objective of the deterministic RCPSP/max scheduling problem is to find a feasible schedule so that the project makespan is minimized.

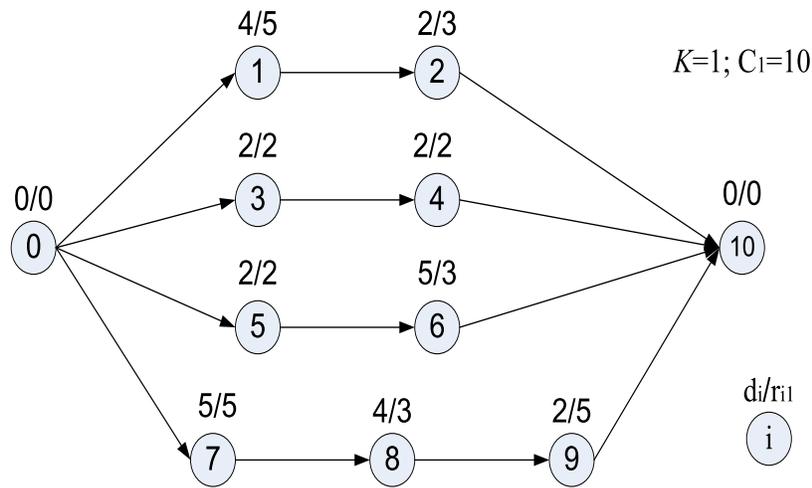

Figure 1: Project Instance.

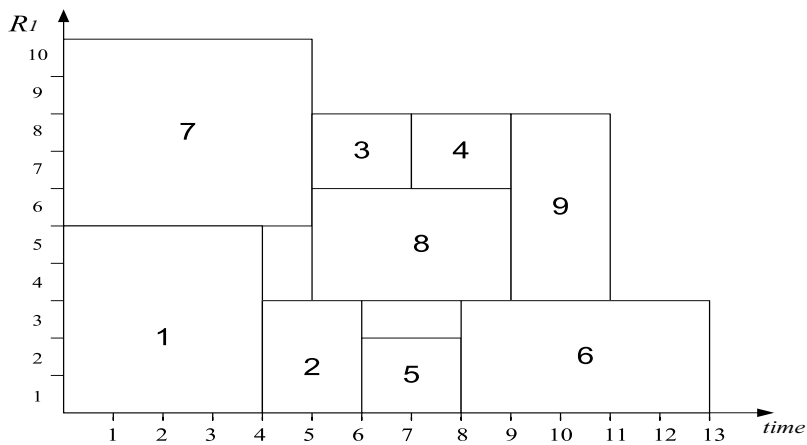

Figure 2: Example Schedule.

**Example 1.** *In Figure 1, we show a simple example of a deterministic RCPSP problem which is a special case of RCPSP/max with only precedence constraints (rather than arbitrary time lags) between activities for expository purposes. Each circle indicates an activity with the number inside the circle representing the activity ID. The two numbers on top of each activity represent the duration and the number of units of the resource required by the activity. In this example, there are 9 activities and one type of resource, with the capacity of the resource limited to 10. It should be noted that the activities 0 and 10 are dummy activities introduced to have a source and sink in the dependency graph. Arrows between activities represent temporal dependencies. A feasible schedule with makespan of 13 is represented in Figure 2.*

### 2.2.2 RCPSP/max with Durational Uncertainty and Robust Makespan

In this paper, we consider RCPSP/max problems with durational uncertainty. The duration of an activity is specified as a sum of its mean value and its deviation: $\tilde{d}_i = d_i^0 + \tilde{z}_i$, where $d_i^0$ is the mean of $\tilde{d}_i$ and $\tilde{z}_i$ is the perturbation part with an expected value of 0 and standard deviation $\sigma$. It should be noted that irrespective of its distribution type, we can always represent $\tilde{d}_i$ as $\tilde{d}_i = d_i^0 + \tilde{z}_i$ where $d_i^0$ is the mean and $\tilde{z}_i$ is the perturbation part with $E(\tilde{z}_i) = 0$. In addition, we also assume that these random variables, $\{\tilde{z}_i\}$, corresponding to durational uncertainty are independent of each other.

Similar to the deterministic RCPSP/max, the *start-to-start* constraints are still deterministic. However, unlike the deterministic case, other types of constraints (*end-to-start* etc.) cannot be converted into deterministic *start-to-start* constraints . Instead the equivalent *start-to-start* constraint is a stochastic one as shown in the following expressions for an *end-to-start* constraint. It should be noted that even though the converted constraints are stochastic, our techniques will still be applicable (with minor modifications) to all types of time lag constraints. Our robust local search techniques depend on the computation of maximum and sum of random variables and even with stochastic time lag constraints that remains the case. In this paper, for purposes of exposition, we present our techniques assuming the temporal dependencies are provided as *start-to-start* constraints.

$$st(a_j) - et(a_i) \leq T_{ij}^{max}$$
$$st(a_j) - (st(a_i) + \tilde{d}_i) \leq T_{ij}^{max}$$
$$st(a_j) - st(a_i) \leq T_{ij}^{max} + \tilde{d}_i$$

In the deterministic setting, start time schedules can be computed and values of makespan can be used to evaluate the performance of the schedule. However, when durational uncertainty is involved, the project makespan becomes a random variable and the schedule is replaced by an execution strategy. In the following sections, we introduce the *Partial Order Schedule* (POS) (Policella et al., 2004), which serves as an execution strategy of the scheduling project.

Given a level of risk $0 < \varepsilon \leq 1$, the goal of our problem is to find such a strategy with a minimum value (across all strategies) of the robust makespan. We define the *robust makespan* as a makespan value where the probability that any feasible schedule (i.e. an assignment of start times to activities) instantiated from the strategy can be completed before robust makespan is at least $1 - \varepsilon$.

### 2.2.3 Partial Order Schedule

A *Partial Order Schedule* (POS) was first proposed by Policella et al. (2004). It is defined as a set of activities, which are partially ordered such that any schedule with total activity order that is consistent with the partial order is resource and time feasible. Mathematically, a POS can be represented by a graph where a node represents an activity and the edges represent the precedence constraints between the activities. Within a POS, each activity retains a set of feasible start times, which provide the flexibility to respond to unexpected disruptions. A POS can be constructed from a given RCPSP instance via a chaining algorithm (where one such algorithm is described below).

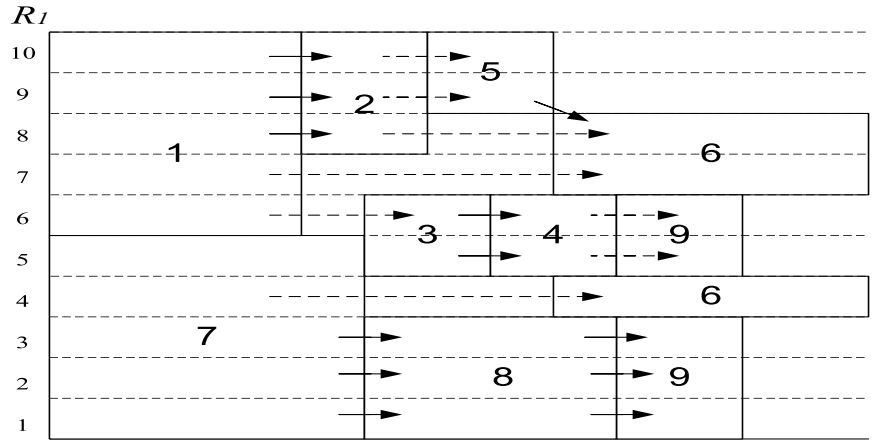

Figure 3: Example of POS

**Example 2.** *Figure 3 provides a POS for the problem instance introduced in Example 1. There are 10 units of the resource and that is shown on the left most side of the figure. Each unit represents a chain. An activity can require multiple resource units and hence is shown on multiple resource units. For instance, activity 6 is shown on resource units 4, 7 and 8. A solid arrow between activities represents a temporal dependency provided in the original problem. Solid arrow between activities 1 and 2 is one such example. A dotted arrow between activities represents a temporal dependency that is introduced since both activities have to be executed on the same resource unit. It is added to remove resource conflict. An example for this is the dependency introduced between activity 2 and activity 6. For explanatory purposes we only consider one resource type in this example, however in the most general case, there exists multiple resource types and a dependency diagram for every resource type.*

### 2.2.4 Chaining Algorithm

Chaining is a procedure of dispatching activities to different resource units (henceforth referred to as chains) based on temporal and resource feasibility. During the chaining process, each activity can be allocated to one or more resource chains based on the number of resource requirement of the activity. During the chaining process, once an activity is scheduled to be executed on a resource unit, an additional edge (indicating precedence relationship) is added between the last activity of the selected chain and this activity so as to eliminate all possible resource conflicts.

In the following, we describe the basic chaining algorithm proposed by Policella et al. (2004). In this algorithm, a feasible schedule is first obtained using a simple greedy heuristic. Consequently, the POS is constructed through a chaining method as follows: First, the set of activities are sorted according to their start times given in the feasible solution; Then, all activities are allocated to different chains in that order, where each chain corresponds to a unit of a certain type of resource. A chain is called *available* for an activity if the end time of the last activity allocated on this chain is no greater than the start time of the activity in the feasible schedule. Once an activity is allocated on a chain, a precedence constraint between this activity and the last activity of the chain is posted. For those activities that require more than one unit of one or more types of resources, they will be allocated to a number of chains with the number equal to the overall number of resource units required by the activity.

**Example 3.** *Take Figure 3 for example. Given the schedule of Figure 2 as an input,activities are first sorted according to their starting time and the sequence of activities can be presented as: (7,1,2,8,3,5,4,6,9). The chaining procedure first picks activity 7 and randomly allocates it to five chains to fulfill its resource requirement. The available chains are those belonging to dummy activity 0.Thus, five chains 1 through 5 are created which posts the precedence relationship from the current last activity 0 to activity 7. Activity 7 then becomes the last activity on those chains. Activity 1 is treated in the same way. The available chains for activity 2 are those belonging to activity 1. Activity 2 is then randomly assigned to chain 8 through 10 and an edge between activity 1 and activity 2 indicating precedence relationship is added. This procedure continues until all activities are dispatched to chains that the number equals its resource requirement, and finally the chained POS 3 is yielded. However, because the randomness of the chaining procedure, activity 6 is allocated to chains that belong to three different activities: activity 2, activity 1 and activity 7. This will tie together the execution of three previously unrelated activities: (activity 2, activity 6),(activity 1, activity 6) and (activity 7, activity 6), which would decrease the flexibility of execution.*

To reduce inter-dependencies between activities as much as possible during the chaining procedure, Policella, Cesta, Oddi, and Smith (2009) developed two heuristics. One direct advantage of such approaches is that synchronization points of a solution can be reduced:

- Activities that require more than one resource units are allocated to the same subset of chains. This is achieved by scanning the list of available chains where the last activity in the chain : (a) requires multiple resource units; and (b) was also previously assigned another resource unit allocated to the current activity.

- Activities with a precedence constraint defined in the original problem are allocated to the same set of chains. This is implemented by choosing a chain that has a last activity with precedence constraint with the current activity.

**Example 4.** *Figure 4 provides the POS computed by using the above mentioned chaining algorithm for the RCPSP problem described in Example 1. When allocating activity 6,*

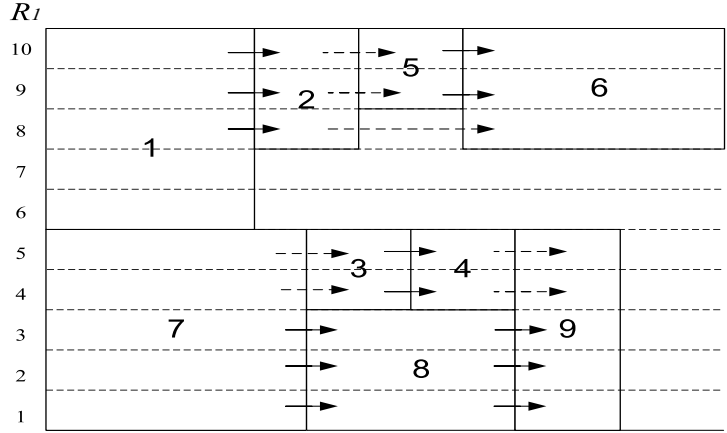

Figure 4: POS computed with Removed Synchronization Point

*the available chains are divided into two sets: {chain 10, chain 9} and {chain 8, chain7, chain6}. The first set contains chains for which the last activity (i.e. activity 5) is already ordered in problem definition with respect to activity 6. A chain (for example, chain 10) is randomly chosen from this set with the last activity on it as activity 5. Then, The remaining available chains for activity 6 is redivided into two sets: {chain 9} and {chain 8, chain7, chain6}. The first set contains the chains with activity 5 (i.e. the last activity of the first picked chain) as the last activity and the second set are the remaining. Activity 6 is first allocated to chains belonging to the first subset to satisfy all remaining resource requirements. In this case, the synchronization points caused by activities 1 and 6, activities 7 and 6 being allocated to different chains has disappeared.*

## 2.3 Job-shop Scheduling Problem (JSP) with Durational Uncertainty

The classical JSP consists of a set of $n$ jobs and a set of $M$ machines. Each job $J_i$ $(i = 1, ...n)$ consists of a sequence of $n_i$ operations denoted as $O_{ij}$ $(j = 1, ...n_i)$ which have to be processed in a given order. For convenience, we enumerate all operations of all jobs by $O_k$, where $k = 1, ...N$ and $N = \sum_{j=1}^{n} n_j$. Each operation $O_k$ has a positive duration denoted as $d_k$ and must be executed on a dedicated machine denoted as $M_k$. Once an operation is started it must be executed for its entire duration. No operations that require the same resource can overlap in their execution. Thus, operations can be partitioned into two sets: job sets and resource sets. Job sets referring to operations corresponding to a job and resource sets referring to all operations that require the same resource.

A solution $s$ is a total ordering of operations on each resource set, which does not conflict with the job ordering. A *path* of a solution $s$ is a sequence of operations which follows both the job ordering and the ordering on various resource sets of the solution $s$. The *length* of a path is equal to the sum of the durations of the operations in the path. The *makespan* of a solution $s$ *make(s)* is the length of the longest path. *The minimum makespan* of a JSP problem is defined to be the minimum value of makespans over all solutions, i.e. $min_s$ *make(s)*. Each operation $O_k$ is associated with a start time of $st(O_k)$ and end time of

51

$et(O_k)$. A *schedule* is an assignment of starting times $st(O_k)$ ($k = 1, ...N$) to all operations on the machines. The objective is to find a schedule which optimizes the total makespan (makespan is the completion time of the last operation): $max_{k=1}^N et(O_k)$, which is also the minimum value of the longest path of all solutions. The job shop scheduling problem is a special case of RCPSP in which resources have unary capacity and each activity (i.e. operation) consumes only one resource.

We can propagate the same notations from RCPSP/max with durational uncertainty to the JSP with durational uncertainty, i.e. the processing time of each activity (i.e. operation) $\tilde{d}_{O_k}$ is now modeled as a sum of an expected value $d_{O_k}^0$ and a random part $\tilde{z}_{O_k}$: $\tilde{d}_{O_k} = d_{O_k}^0 + \tilde{z}_{O_k}$. The objective is to find the robust makespan with a given level of risk.

## 2.4 Segregated Random Variables

A *primitive* random variable $\tilde{z}_k$ is one which has zero mean. Examples of a primitive random variable include $U(-a, a)$ (uniform distribution between constants $-a$ and $a$) and $N(0, \sigma)$ (normal distribution with mean 0 and variance $\sigma^2$). As mentioned earlier, we assume that every uncertain distribution is equal to the sum of its nominal value (mean) and its deviation, represented by one (or possibly more) primitive random variable $\tilde{z}$. In a straight forward representation, there is only one primitive random variable $\tilde{z}_k$ associated with an uncertain variable. In the recent work by Chen, Sim, Sun, and Zhang (2008), each primitive random variable $\tilde{z}_k$ is represented by 2 *segregated* random variables $\tilde{z}_k^+$ (read z-plus) and $\tilde{z}_k^-$ (z-minus):

$$\tilde{z} = \tilde{z}^+ - \tilde{z}^- \tag{6}$$

$$\tilde{z}^+ = max \{\tilde{z}, 0\} \tag{7}$$

$$\tilde{z}^- = max \{-\tilde{z}, 0\} . \tag{8}$$

In the following Table 1, we give examples of the respective values of mean $\mu_p$, $\mu_m$ and variance $\sigma_p^2$, $\sigma_m^2$ for the segregated variables $\tilde{z}^+$ and $\tilde{z}^-$.

| $\tilde{z}$ | $Var(\tilde{z})$ | $\sigma_p^2$, $\sigma_m^2$ | $\mu_p, \mu_m$ |
|---|---|---|---|
| $U(-a, a)$ | $\frac{a^2}{3}$ | $\frac{5a^2}{48}$ | $\frac{a}{4}$ |
| $N(0, \sigma)$ | $\sigma^2$ | $\frac{(\pi-1)\sigma^2}{2\pi}$ | $\frac{\sigma}{\sqrt{2\pi}}$ |

Table 1: Values of the mean and variance for the segregated variables under Uniform and Normal Distribution

The underlying assumption with the use of segregated random variables is that the mean and variance of the individual segregated variables is provided for the random variables employed. We are not aware of mean and variance values for segregated variables for distributions other than normal and uniform.

## 2.5 Decision Rules for Optimization under Data Uncertainty

In optimization problems with data uncertainty, a decision rule specifies the dependence of adjustable variables on the uncertainty parameters and the non-adjustable variables. Let $\tilde{\mathbf{z}}$ and $\mathbf{x}$ denote the set of primitive random variables and non-adjustable variables respectively. An example is the linear decision rule framework proposed by Ben-Tal and Nemirovski (2002), where the setting value of an adjustable decision variable $\tilde{S}(\mathbf{x}, \tilde{\mathbf{z}})$ is assumed to be affinely dependent on a subset of the $N$ number of primitive random variables:

$$\tilde{S}(\mathbf{x}, \tilde{\mathbf{z}}) = c^0 + \sum_{k=1}^{N} c_k(\mathbf{x}) \tilde{z}_k \qquad (9)$$

where each $c_k(\mathbf{x})$ $(1 \leq k \leq N)$ is a coefficient derived from $\mathbf{x}$.

Another example is the segregated linear decision rule framework proposed by Chen et al. (2008), where each adjustable decision variable is assumed to be affinely dependent on a set of some $N$ segregated random variables $\{\tilde{z}_1^+, \tilde{z}_1^-, \ldots, \tilde{z}_N^+, \tilde{z}_N^-\}$. Hence, a segregated linear decision rule has the following general form:

$$\tilde{S}(\mathbf{x}, \tilde{\mathbf{z}}) \quad = c^0 + \sum_{k=1}^{N} \left\{ c_k^+ \tilde{z}_k^+ + c_k^- \tilde{z}_k^- \right\}. \qquad (10)$$

As we will show below, a segregated linear decision rule allows us to easily obtain an upper bound on a subset of random variables (see Eqn 14), which is not possible in the linear decision rule framework proposed by Ben-Tal and Nemirovski (2002).

Given the mean and variance for each segregated variable $E(\tilde{z}_k^+) = E(\tilde{z}_k^-) = \mu_k$, $Var(\tilde{z}_k^+) = \sigma_{pk}^2$ and $Var(\tilde{z}_k^-) = \sigma_{mk}^2$, we can express the expected value and variance of any adjustable variable as:

$$E[\tilde{S}(\mathbf{x}, \tilde{\mathbf{z}})] = c^0 + \sum_{k=1}^{N} \left\{ c_k^+ \mu_k + c_k^- \mu_k \right\} \qquad (11)$$

$$Var[\tilde{S}(\mathbf{x}, \tilde{\mathbf{z}})] = \sum_{k=1}^{N} \left\{ \left[ c_k^+ \sigma_{pk} \right]^2 + \left[ c_k^- \sigma_{mk} \right]^2 - 2 c_k^+ c_k^- \mu_k \right\}. \qquad (12)$$

## 3. Decision Rules for RCPSP/max with Durational Uncertainty

In RCPSP/max with durational uncertainty, a decision rule specifies the dependence of activity start times on the durational uncertainty associated with other activities. To make the comparison with Equation 9, $\mathbf{x}$ represents the POS to be generated; each task's start time is associated with the adjustable variable $\tilde{S}(\mathbf{x}, \tilde{\mathbf{z}})$, where $c^0$ represents the earliest start time of this task under the POS, and $c_k(\mathbf{x})$ encodes how task $k$ is related to this task in the POS.

In a scheduling context, the start time of an activity is dependent on the start times of the preceding activities, i.e. Adjustable variables $\tilde{S}(\mathbf{x}, \tilde{\mathbf{z}})$ are dependent on one another. Any activity will either start after the end of an activity (i.e. in series) or after the end of multiple activities occurring simultaneously (i.e. in parallel). Thus, adjustable variables are functions of other adjustable variables through the addition operator (to model serial activities) and/or the maximum operator (to model parallel activities).

Given $M$ number of adjustable variables, we may express its sum as an adjustable variable in the form of a segregated linear decision rule as follows:

$$\sum_{i=1}^{M} \tilde{S}_i(\mathbf{x}, \tilde{\mathbf{z}}) = \sum_{i=1}^{M} c_i^0 + \sum_{k=1}^{N} \left\{ \sum_{i=1}^{M} c_{i,k}^+ \tilde{z}_k^+ + \sum_{i=1}^{M} c_{i,k}^- \tilde{z}_k^- \right\}. \tag{13}$$

Similarly, given some set $C$ of adjustable variables, we may also express the upper bound on the maximum of these variables as an adjustable variable in the form of a segregated linear decision rule:

$$\max_{i \in C} \{\tilde{S}_i(\mathbf{x}, \tilde{\mathbf{z}})\} \leq \max_{i \in C} \{c_i^0\} + \sum_{k=1}^{N} \left\{ \max_{i \in C} \{c_{i,k}^+\} \tilde{z}_k^+ \right\} + \sum_{k=1}^{N} \left\{ \max_{i \in C} \{c_{i,k}^-\} \tilde{z}_k^- \right\}. \tag{14}$$

More specifically, the output of solving a RCPSP/max involves a POS that is represented as a graph with activities as vertices and precedence constraints between activities as the edges. Given a POS graph, $\mathbf{x} = (V, E)$, where $V$ is the set of activities and $E$ is the set of temporal dependencies (an edge $(u, v)$ represents a temporal dependency that states that activity $v$ should occur after activity $u$). For any activity $v \in V$, the decision rule for computing its start time is defined recursively as follows:

$$\tilde{S}_v(\mathbf{x}, \tilde{\mathbf{z}}) = \max_{(u,v) \in E} \{d_u^0 + \tilde{z}_u + \tilde{S}_u(\mathbf{x}, \tilde{\mathbf{z}})\}. \tag{15}$$

Equation 15 is a recursive expression that is defined as a combination of *sum* and *maximum* on a set of random variables. It should be noted that combinations of *sum* and *maximum* of random variables cannot be computed exactly and hence we present two *operational* decision rule approximations to evaluate the recursive expression of Equation 15: (a) Segregated Linear Approximation(SLA); and (b) General Non-Linear Approximation(GNLA). It should be noted that the $\tilde{S}_v$ is computable as long as mean and variance of $\tilde{S}_u$ is computable and this is demonstrated with both our approximations.

## 3.1 Segregated Linear Approximation (SLA)

In this decision rule, the duration for each activity is defined based on the segregated random variables introduced in Section 2.4. For an uncertain duration $\tilde{d}$ with mean processing time $d^0$, we represent $\tilde{d}$ as a sum of three components: its mean $d^0$, lateness $\tilde{z}^+$ (i.e. $max\{\tilde{d} - d^0, 0\}$), and earliness $\tilde{z}^-$ (i.e. $max\{d^0 - \tilde{d}, 0\}$),

$$\tilde{d} = d^0 + \tilde{z}^+ - \tilde{z}^-. \tag{16}$$

For a normally distributed duration, i.e., $\tilde{z} \sim N\{0, \sigma\}$, the respective values of mean and variance for the segregated variables can be summarized as:

$$E[\tilde{z}^+] = E[\tilde{z}^-] = \frac{\sigma}{\sqrt{2\pi}} \tag{17}$$

$$Var[\tilde{z}^+] = Var[\tilde{z}^-] = \frac{(\pi - 1)\sigma^2}{2\pi}. \tag{18}$$

Now we describe the computation of $\tilde{S}_v(\mathbf{x}, \tilde{\mathbf{z}})$ by representing durational uncertainty for activities using segregated random variables. Upper bounds on both the sum and maximum of random variables are derived as linear functions of segregated variables as illustrated below:

- *Sum* of random variables : In the case of a project network involving $k$ activities, any two of which have either precedence constraints in between or competing for the same resource units, a solution in the form of POS requires computation of the *sum* of activity durations. The start time of the activity starting after the $k$-activity project is expressed as:

$$\tilde{S}_k(\mathbf{x}, (\tilde{\mathbf{z}}^+, \tilde{\mathbf{z}}^-)) \quad = \sum_{i=1}^{k}(d_i^0 + \tilde{z}_i^+ - \tilde{z}_i^-). \tag{19}$$

  Thus, the adjustable variable $\tilde{S}_k$ a mean of $\sum_{i=1}^{k} d_i^0$ with uncertainty captured by a random variable, which has a positive segregated component of $\sum_{i=1}^{k} \tilde{z}_i^+$ and a negative segregated component of $\sum_{i=1}^{k} \tilde{z}_i^-$. Mean and variance of the segregated variables are known and hence the mean and variance of $\tilde{S}_k$ are easy to compute.

- *Max* of random variables: Consider activities that are executed concurrently, the upper bound on the start time of an activity starting after the parallel $k$-activity project network in SLA is represented by a linear function of the positive segregated components of duration perturbations:

$$\tilde{S}_k(\mathbf{x}, (\tilde{\mathbf{z}}^+, \tilde{\mathbf{z}}^-)) \quad \leq \max_{i=1,\ldots k}\{d_i^0\} + \sum_{i=1}^{k} \tilde{z}_i^+. \tag{20}$$

  Thus, the adjustable variable $\tilde{S}_k$ has an upper bound on the mean of $\max_{i=1,\ldots k}\{d_i^0\}$ with uncertainty captured by a random variable with the positive segregated component given by $\sum_{i=1}^{k} \tilde{z}_i^+$ and no negative segregated component. Mean and variance of the segregated variables are known and hence the mean and variance of $\tilde{S}_k$ are easy to compute.

Since, in both cases (*sum* and *max*) $\tilde{S}_k$ is expressed linearly on a subset of random segregated variables, the recursive computation is straightforward. Compared with other linear decision rules (Ben-Tal & Nemirovski, 2002), the superiority of SLA (Chen et al., 2008) lies in this ability to linearly express an upper bound on a subset of random variables by dissecting each uncertainty into its positive and negative components. While this approximation increases tractability and scalability, it comes at the expense of losing accuracy.

### 3.2 General Non Linear Approximation (GNLA)

While SLA is efficient, it can typically provide loose upper bounds on robust makespan due to the linear approximation for computing max of random variables. In this section, we describe General Non Linear Approximation (GNLA), which is not restricted to only affine dependencies. For clarity and comparison purposes, we use $\tilde{G}$ to denote the start time instead of $\tilde{S}$ used in SLA.

Given the mean and variance values of duration uncertainty, we describe the approximation involved in computing mean and variance of the *sum* and *max* of activities that will be used in Equation 15. It should be recalled that irrespective of the distribution of the uncertain duration $\tilde{d}$, we can always represent $\tilde{d}$ as $\tilde{d} = d^0 + \tilde{z}$, where $d^0$ is the mean of $\tilde{d}$ and $\tilde{z}$ is the perturbation part. Thus, $E(\tilde{z}) = 0$.

### 3.2.1 SUM OF RANDOM VARIABLES

We compute sum of all stochastic durations in a serial $k$ activity project network as follows:

$$\tilde{G}_k(\mathbf{x}, \tilde{\mathbf{z}}) = \sum_{i=1}^{k} (d_i^0 + \tilde{z}_i). \tag{21}$$

In this case, we have a similar representation to SLA. Mean and variance of $\tilde{G}_k$ are computed as follows:

Since $\{\tilde{z}_i\}_{i=1,\ldots k}$ are random variables with zero mean, we can then calculate the expected value as:

$$E[\sum_{i=1}^{k} (d_i^0 + \tilde{z}_i)] = \sum_{i=1}^{k} d_i^0. \tag{22}$$

Because $\{\tilde{z}_i\}$ are assumed to be independent of each other, the variance value is computed by the following expression:

$$Var[\sum_{i=1}^{k} (d_i^0 + \tilde{z}_i)] = \sum_{i=1}^{k} Var[\tilde{z}_i], \tag{23}$$

and under normal distribution where $\tilde{z}_i \sim N(0, \sigma_i)$, we have

$$Var[\sum_{i=1}^{k} (d_i^0 + \tilde{z}_i)] = \sum_{i=1}^{k} \sigma_i^2. \tag{24}$$

Note that the expressions for expected value and variance in the case of serial activities are identical to the ones used by Wu, Brown, and Beck (2009).

### 3.2.2 MAX OF RANDOM VARIABLES

For ease of explanation, we begin by considering two activities to be executed in parallel and then extend the analysis to multiple parallel activities. In GNLA, (unlike in SLA) the *max* of random variables itself is not approximated but the expected value and variance of the *max* are approximately calculated.

***Expected Value and Variance of* Max *of Two Variables***

The decision rule to represent the starting time of an activity, which will begin after the completion of two parallel activities, is defined as:

$$\tilde{G}_2(\tilde{\mathbf{z}}) \leq max\{d_1^0, d_2^0\} + max\{\tilde{z}_1, \tilde{z}_2\}. \tag{25}$$

Note that we tighten the bound in Eqn 20 by replacing $\tilde{z}_1^+ + \tilde{z}_2^+$ with $max\{\tilde{z}_1, \tilde{z}_2\}$.

We now derive the expressions for expected value and variance of the adjustable variable, i.e., the RHS term of Eqn 25. Firstly, we focus on the expected value:

$$E[max\{d_1^0, d_2^0\} + max\{\tilde{z}_1, \tilde{z}_2\}] = max\{d_1^0, d_2^0\} + E[max\{\tilde{z}_1, \tilde{z}_2\}]. \tag{26}$$

In the general case, it is difficult to derive an exact expression for $E[max\{\tilde{z}_1, \tilde{z}_2\}]$ and hence, we provide an upper bound.

In the following Propositions 1 and 2, we compute expected value and variance for the more general case of $E(\tilde{z}) \geq 0$ (note that we assume $E(\tilde{z}) = 0$ for all primitive random variables). We calculate for the more general case because it will be required in the computation of expected value and variance for more than two random variables (next subsection).

**Proposition 1.** *The expected value for the maximum of two general distributions, $\tilde{z}_1$ and $\tilde{z}_2$ with nonnegative means is less than*
$\frac{1}{2}(E[\tilde{z}_1] + E[\tilde{z}_2]) + \frac{1}{2}\sqrt{Var[\tilde{z}_1] + Var[\tilde{z}_2] + (E[\tilde{z}_1])^2 + (E[\tilde{z}_2])^2}.$

**Proof.** We begin by considering the following two equalities:

$$max\{\tilde{z}_1, \tilde{z}_2\} + min\{\tilde{z}_1, \tilde{z}_2\} = \tilde{z}_1 + \tilde{z}_2$$
$$max\{\tilde{z}_1, \tilde{z}_2\} - min\{\tilde{z}_1, \tilde{z}_2\} = |\tilde{z}_1 - \tilde{z}_2|.$$

We now sum the above two equalities.

$$max\{\tilde{z}_1, \tilde{z}_2\} = \frac{1}{2}(\tilde{z}_1 + \tilde{z}_2 + |\tilde{z}_1 - \tilde{z}_2|). \tag{27}$$

Thus, we can now compute the expected value of the maximum using the following equation:

$$E[max\{\tilde{z}_1, \tilde{z}_2\}] = \frac{1}{2}(E[\tilde{z}_1] + E[\tilde{z}_2] + E|\tilde{z}_1 - \tilde{z}_2|). \tag{28}$$

In addition, by using the definition of variance, we obtain:

$$Var|\tilde{z}_1 - \tilde{z}_2| = E(\tilde{z}_1 - \tilde{z}_2)^2 - (E|\tilde{z}_1 - \tilde{z}_2|)^2 \geq 0.$$

Therefore,

$$\begin{aligned} E|\tilde{z}_1 - \tilde{z}_2| &\leq \sqrt{E(\tilde{z}_1 - \tilde{z}_2)^2} \\ &= \sqrt{E(\tilde{z}_1^2) + E(\tilde{z}_2^2) - 2E(\tilde{z}_1)E(\tilde{z}_2)} \\ &\leq \sqrt{E(\tilde{z}_1^2) + E(\tilde{z}_2^2)} \\ &= \sqrt{Var[\tilde{z}_1] + Var[\tilde{z}_2] + E(\tilde{z}_1)^2 + E(\tilde{z}_2)^2}. \end{aligned} \tag{29}$$

Substituting the final expression of Eqn 29 into Eqn 28 yields the bound

$$E[max\{\tilde{z}_1, \tilde{z}_2\}] \leq \frac{1}{2}(E[\tilde{z}_1] + E[\tilde{z}_2]) + \frac{1}{2}\sqrt{Var[\tilde{z}_1] + Var[\tilde{z}_2] + (E[\tilde{z}_1])^2 + (E[\tilde{z}_2])^2}. \tag{30}$$

Hence the proof.

Note that in this paper, we assume $E(\tilde{z}) = 0$, thus, a tighter bound can be obtained from Eqn 30:

$$E[max\{\tilde{z}_1, \tilde{z}_2\}] \leq \frac{1}{2}\sqrt{Var[\tilde{z}_1] + Var[\tilde{z}_2]}. \tag{31}$$

In the special case where $\{\tilde{z}_i\}$ $(i = 1, ...k)$ are normally and identically distributed, i.e. $\tilde{z}_i \sim N(0, \sigma)$, we know from the work of Clark (1961) that there is a closed form representation for the expected value of the maximum when $k = 2$:

$$E[max\{\tilde{z}_1, \tilde{z}_2\}] = \frac{\sigma}{\sqrt{\pi}}.$$

Now we focus on deriving expressions for variance of the maximum of two general distributions, i.e., $Var[max(\tilde{z}_1, \tilde{z}_2)]$.

**Proposition 2.** *The variance for the maximum of two general distributions, $\tilde{z}_1$ and $\tilde{z}_2$ with nonnegative means is less than $Var(\tilde{z}_1) + Var(\tilde{z}_2) + \frac{1}{2}(E(\tilde{z}_1))^2 + \frac{1}{2}(E(\tilde{z}_2))^2$.*

**Proof.** From Eqn 27, we have

$$\begin{aligned}
Var[max(\tilde{z}_1, \tilde{z}_2)] &= \tfrac{1}{4}Var[\tilde{z}_1 + \tilde{z}_2 + |\tilde{z}_1 - \tilde{z}_2|] \\
&= \tfrac{1}{4}(Var[\tilde{z}_1 + \tilde{z}_2] + Var|\tilde{z}_1 - \tilde{z}_2| + 2COV(\tilde{z}_1 + \tilde{z}_2, |\tilde{z}_1 - \tilde{z}_2|)) \\
&\leq \tfrac{1}{4}(Var[\tilde{z}_1 + \tilde{z}_2] + Var|\tilde{z}_1 - \tilde{z}_2| + 2\sqrt{Var[\tilde{z}_1 + \tilde{z}_2]Var|\tilde{z}_1 - \tilde{z}_2|}) \\
&\leq \tfrac{1}{2}(Var[\tilde{z}_1 + \tilde{z}_2] + Var|\tilde{z}_1 - \tilde{z}_2|).
\end{aligned} \tag{32}$$

Firstly, we consider the following two equations.

$$\begin{aligned}
Var|\tilde{z}_1 - \tilde{z}_2| &= E(\tilde{z}_1 - \tilde{z}_2)^2 - (E|\tilde{z}_1 - \tilde{z}_2|)^2 \\
Var(\tilde{z}_1 - \tilde{z}_2) &= E(\tilde{z}_1 - \tilde{z}_2)^2 - (E(\tilde{z}_1 - \tilde{z}_2))^2
\end{aligned} \tag{33}$$

Subtracting the second from the first yields

$$Var|\tilde{z}_1 - \tilde{z}_2| = Var(\tilde{z}_1 - \tilde{z}_2) + (E(\tilde{z}_1 - \tilde{z}_2))^2 - (E|\tilde{z}_1 - \tilde{z}_2|)^2.$$

Now, we substitute this expression into the last term of Eqn 32 to obtain:

$$Var[max(\tilde{z}_1, \tilde{z}_2)] \leq Var(\tilde{z}_1) + Var(\tilde{z}_2) + \tfrac{1}{2}(E(\tilde{z}_1) - E(\tilde{z}_2))^2 - \tfrac{1}{2}(E|\tilde{z}_1 - \tilde{z}_2|)^2. \tag{34}$$

When no specific distribution about duration perturbation is known, we can obtain a bound for $Var[max(\tilde{z}_1, \tilde{z}_2)]$ as:

$$Var[max(\tilde{z}_1, \tilde{z}_2)] \leq Var(\tilde{z}_1) + Var(\tilde{z}_2) + \tfrac{1}{2}(E(\tilde{z}_1))^2 + \tfrac{1}{2}(E(\tilde{z}_2))^2. \tag{35}$$

Hence the proof.

Note that in this paper, we assume $E(\tilde{z}) = 0$, thus, a tighter bound can be obtained from Eqn 35:

$$Var[max(\tilde{z}_1, \tilde{z}_2)] \leq Var(\tilde{z}_1) + Var(\tilde{z}_2). \tag{36}$$

It is interesting to consider the special case when both random variables are normally distributed. We first state the following lemma[1].

---

1. This can be found in statistics texts, and found online at http://en.wikipedia.org/wiki/Half-normal_distribution.

**Lemma 3.1.** *If $X$ is normally distributed $X \sim N(0, \sigma)$, then $Y = |X|$ is half-normally distributed, with*

$$E(Y) = \sigma\sqrt{\frac{2}{\pi}}. \tag{37}$$

Under normal distribution $\tilde{z}_i \sim N(0, \sigma_i)$, since $\tilde{z}_1 - \tilde{z}_2$ is also normally distributed, and $\tilde{z}_1 - \tilde{z}_2 \sim N(0, \sigma_1 + \sigma_2)$, we can conclude from Lemma 3.1 that $|\tilde{z}_1 - \tilde{z}_2|$ follows half-normal distribution with

$$E|\tilde{z}_1 - \tilde{z}_2| = (\sigma_1 + \sigma_2)\sqrt{\frac{2}{\pi}}. \tag{38}$$

Thus, if we substitute this expression into Eqn 34, we can express an upper bound on the variance value for the maximum duration perturbation of two activities, when $\tilde{z}_i \sim N(0, \sigma_i)$ as :

$$Var[max(\tilde{z}_1, \tilde{z}_2)] \leq (1 - \frac{1}{\pi})(\sigma_1^2 + \sigma_2^2) - \frac{2}{\pi}\sigma_1\sigma_2. \tag{39}$$

**Expected Value and Variance of *Max* of Multiple Variables**

Extending from two to $k$ ($k > 2$) parallel activities, the completion time can be upper bounded by:

$$\tilde{G}_k(\tilde{\mathbf{z}}) \leq \max_{i=1,\ldots k}\{d_i^0\} + \max_{i=1,\ldots k}\{\tilde{z}_i\}. \tag{40}$$

In the following, we first compute the variance value of the above RHS term and then use a similar procedure to compute the expected value. The basic expression for variance of RHS is:

$$Var[\max_{i=1,\ldots k}\{d_i^0\} + \max_{i=1,\ldots k}\{\tilde{z}_i\}] = Var[\max_{i=1,\ldots k}\{\tilde{z}_i\}]. \tag{41}$$

To obtain the value of $Var[\max_{i=1,\ldots k}\{\tilde{z}_i\}]$ for general probability distributions, we take advantage of the analysis provided for the two-parallel-activity case above. The following steps outline the overall idea:

(a) Firstly, we group the activity set $\{a_1, ..., a_k\}$ into a couple set $\{C_1, ..., C_{\lceil\frac{k}{2}\rceil}\}$, where each element $C_j(j = 1, ... \lceil\frac{k}{2}\rceil)$ contains two different activities $C_j = \{a_{j1}, a_{j2}\}$ chosen from the activity set. Note that when $k$ is an odd, the final element in the couple set contains just one activity.

(b) For each couple $C_j$, we apply the maximum operator on duration perturbations of involving activities. Denote $\tilde{c}_j = max\{\tilde{z}_{j1}, \tilde{z}_{j2}\}$, where $\tilde{z}_{j1}$ and $\tilde{z}_{j2}$ are duration perturbations of the two activities involved in $C_j$, then $Var(\tilde{c}_j)$ can be calculated based on the expression for the two-parallel-activity case.

(c) Then we have $\max_{i=1,\ldots k}\{\tilde{z}_i\} = \max_{j=1,\ldots\lceil\frac{k}{2}\rceil}\{\tilde{c}_j\}$. (Note again just one activity is contained in $C_{\lceil\frac{k}{2}\rceil}$ when $k$ is odd). Then, we can build another couple set from $\{C_1, ..., C_{\lceil\frac{k}{2}\rceil}\}$, and the same method from steps (1) and (2) above is used to compute $Var[\max_{j=1,\ldots\lceil\frac{k}{2}\rceil}\{\tilde{c}_j\}]$ based on Eqn 35 and/or Eqn 36 and/or Eqn 39.

There are numerous ways (exponential in $k$) for generating the couple set $\{C_1, ..., C_{\lceil \frac{k}{2} \rceil}\}$ for $k$ activities in parallel. Each of these couple sets can lead to different levels of tightness of derived robust makespan. To compute the grouping which provides the best robust fitness for random variables with generic distributions is an open problem. Instead, we focus on a heuristic that computes the best grouping under normal distribution $\tilde{z}_i \sim N(0, \sigma_i)$. It is obtained by solving the following optimization problem:

$$\max_t \sum_{j=1,...\lfloor \frac{k}{2} \rfloor} \sigma_{j1} \sigma_{j2} \tag{42}$$

where $t$ denotes the grouping technique and is also the decision variable; $\{C_j\}$ is the couple set constructed from the activity set under grouping method $t$; $\sigma_{j1}$ and $\sigma_{j2}$ are the standard deviations of data perturbation for durations of activities contained in $C_j$. The intuition for employing this optimization problem is obtained from the Equation 39. It should be noted that computing a tighter bound on variance implies considering the highest possible value of the product of primitive variances. Hence, the reason for employing the optimization problem of Equation 42.

**Proposition 3.** *The solution $t^*$ to the optimization problem of Eqn 42 is obtained by ordering the $k$ activities in a non-increasing order of their variance values and then grouping all two nearest activities according to the order, i.e. $C_j = \{a_{j1}, a_{j2}\}$, where $j = 1, ...\lfloor \frac{k}{2} \rfloor$ and the standard deviations are in the following order:*

$$\sigma_{11} \geq \sigma_{12} \geq \sigma_{21} \geq \sigma_{22} \geq, ...\sigma_{\lfloor \frac{k}{2} \rfloor 1} \geq \sigma_{\lfloor \frac{k}{2} \rfloor 2}. \tag{43}$$

**Proof.** Suppose we have another grouping method $t'$, in which all elements in the couple set are the same as under $t^*$ except two couples [2] where the ordering is different, i.e., $C_m = \{a_{m1}, a_{n2}\}$ and $C_n = \{a_{m2}, a_{n1}\}$ $(m \neq n)$, where $C_m = \{a_{m1}, a_{m2}\}$ and $C_n = \{a_{n1}, a_{n2}\}$ under $t^*$. Without loss of generality, assume $m > n$ and from Eqn 43, we have

$$\sigma_{m1} \geq \sigma_{m2} \geq \sigma_{n1} \geq \sigma_{n2}. \tag{44}$$

Since $t'$ is supposed to provide a solution which is no less ( defined in Eqn 42) than $t^*$, i.e.

$\sigma_{11}\sigma_{12} + ... + \sigma_{m1}\sigma_{n2} + ... + \sigma_{n1}\sigma_{m2} + ... + \sigma_{\lfloor \frac{k}{2} \rfloor 1}\sigma_{\lfloor \frac{k}{2} \rfloor 2}$
$\geq$
$\sigma_{11}\sigma_{12} + ... + \sigma_{m1}\sigma_{m2} + ... + \sigma_{n1}\sigma_{n2} + ... + \sigma_{\lfloor \frac{k}{2} \rfloor 1}\sigma_{\lfloor \frac{k}{2} \rfloor 2}.$
Therefore, we have

$$\sigma_{m1}\sigma_{n2} + \sigma_{n1}\sigma_{m2} \geq \sigma_{m1}\sigma_{m2} + \sigma_{n1}\sigma_{n2},$$

which is equivalent to: $(\sigma_{m1} - \sigma_{n1})(\sigma_{n2} - \sigma_{m2}) \geq 0$.
This contradicts Eqn 44 (except the case where all standard deviations are equal, in which case mixing the order does not affect anything). Thus, there exists no such $t'$ which is different from $t^*$ by at least two couples and has better objective value. The general case that $t'$ has multiple (more than two) couples different from $t^*$ can be easily derived from to this case (and is omitted due to space constraints).

---

2. It should be noted that if there is an ordering change in only one couple, then the method still produces the same solution because within a couple the variance computation does not consider the order.

Hence the proof.

As for analyzing the expected value $E[\max_{i=1,...k}\{\tilde{z}_i\}]$, we apply the same procedure employed to calculate the variance, i.e., based on the group solution returned by the above optimization problem, we first calculate the expected value for each couple and then, get the final bound following Eqn 30 and/or Eqn 31 and/or Eqn 32.

At present, we are unable to show the effectivness of our grouping heuristic (Equation 42) analytically in the most general case. However, we show the intuition behind the grouping heuristic by providing an analytical comparison[3] on an example where there are four activities (normally distributed durations) executed in parallel, i.e. $\tilde{z}_i \sim N(0, \sigma_i)$, and we assume $\sigma_1 \geq \sigma_2 \geq \sigma_3 \geq \sigma_4$ (no loss of generality).

The representation of makespan under our grouping heuristic (denoted as $M_{heu}$) and random grouping (denoted as $M_{ran}$) are, respectively:

$$M_{heu} = max\{d_1^0, d_2^0, d_3^0, d_4^0\} + max\{max\{\tilde{z}_1, \tilde{z}_2\}, max\{\tilde{z}_3, \tilde{z}_4\}\}$$
$$M_{ran} = max\{d_1^0, d_2^0, d_3^0, d_4^0\} + max\{max\{\tilde{z}_1, \tilde{z}_4\}, max\{\tilde{z}_2, \tilde{z}_3\}\}. \tag{45}$$

Let us first examine mean and variance values of $M_{heu}$. From Eqn 31, we have

$$E(max\{\tilde{z}_1, \tilde{z}_2\}) \leq \tfrac{1}{2}\sqrt{\sigma_1^2 + \sigma_2^2}$$
$$E(max\{\tilde{z}_3, \tilde{z}_4\}) \leq \tfrac{1}{2}\sqrt{\sigma_3^2 + \sigma_4^2}. \tag{46}$$

From Eqn 39, we have

$$Var[max\{\tilde{z}_1, \tilde{z}_2\}] \leq (1 - \tfrac{1}{\pi})(\sigma_1^2 + \sigma_2^2) - \tfrac{2}{\pi}\sigma_1\sigma_2$$
$$Var[max\{\tilde{z}_3, \tilde{z}_4\}] \leq (1 - \tfrac{1}{\pi})(\sigma_3^2 + \sigma_4^2) - \tfrac{2}{\pi}\sigma_3\sigma_4. \tag{47}$$

From Eqn 30, Eqn 35, Eqn 46 and Eqn 47, we can obtain bounds of mean and variance values of of $M_{heu}$ are [4]:

$$E(M_{heu}) \leq const + \tfrac{1}{4}(\sqrt{\sigma_1^2 + \sigma_2^2} + \sqrt{\sigma_3^2 + \sigma_4^2}) + \tfrac{1}{2}\sqrt{(\tfrac{5}{4} - \tfrac{1}{\pi})\sum_{i=1}^4 \sigma_i^2 - \tfrac{2}{\pi}(\sigma_1\sigma_2 + \sigma_3\sigma_4)}$$
$$Var(M_{heu}) \leq (\tfrac{9}{8} - \tfrac{1}{\pi})\sum_{i=1}^4 \sigma_i^2 - \tfrac{2}{\pi}(\sigma_1\sigma_2 + \sigma_3\sigma_4). \tag{48}$$

Similarly, mean and variance values of of $M_{ran}$ can also be calculated,

$$E(M_{ran}) \leq const + \tfrac{1}{4}(\sqrt{\sigma_1^2 + \sigma_4^2} + \sqrt{\sigma_2^2 + \sigma_3^2}) + \tfrac{1}{2}\sqrt{(\tfrac{5}{4} - \tfrac{1}{\pi})\sum_{i=1}^4 \sigma_i^2 - \tfrac{2}{\pi}(\sigma_1\sigma_4 + \sigma_2\sigma_3)}$$
$$Var(M_{ran}) \leq (\tfrac{9}{8} - \tfrac{1}{\pi})\sum_{i=1}^4 \sigma_i^2 - \tfrac{2}{\pi}(\sigma_1\sigma_4 + \sigma_2\sigma_3). \tag{49}$$

From Eqn 57, bounds of fitness of $M_{heu}$ (denoted by $Fit_{heu}$) and $M_{ran}$ (denoted by $Fit_{ran}$) can then be respectively represented as a function of RHS of Eqn 48 and Eqn 49. We then examine the difference value between the two bounds, $Fit_{heu} - Fit_{ran}$. Let us first compare the first term of RHS of mean values in Eqn 48 and Eqn 49, since

$$(\sqrt{\sigma_1^2 + \sigma_2^2} + \sqrt{\sigma_3^2 + \sigma_4^2})^2 - (\sqrt{\sigma_1^2 + \sigma_4^2} + \sqrt{\sigma_2^2 + \sigma_3^2})^2$$
$$= 2\sqrt{\sigma_1^2\sigma_3^2 + \sigma_2^2\sigma_4^2 + \sigma_1^2\sigma_4^2 + \sigma_2^2\sigma_3^2} - 2\sqrt{\sigma_1^2\sigma_3^2 + \sigma_2^2\sigma_4^2 + \sigma_1^2\sigma_2^2 + \sigma_3^2\sigma_4^2} \tag{50}$$

---

3. The calculation will use the robust fitness function provided in Definition 57 introduced in Section 4.

4. Note that $const$ in Eqn 48 and Eqn 49 is $max\{d_1^0, d_2^0, d_3^0, d_4^0\}$.

and from Proposition 3, we have

$$\sigma_1\sigma_4 + \sigma_2\sigma_3 \leq \sigma_1\sigma_2 + \sigma_3\sigma_4, \tag{51}$$

thus,

$$\sigma_1^2\sigma_4^2 + \sigma_2^2\sigma_3^2 - (\sigma_1^2\sigma_2^2 + \sigma_3^2\sigma_4^2) = (\sigma_1\sigma_4 + \sigma_2\sigma_3)^2 - (\sigma_1\sigma_2 + \sigma_3\sigma_4)^2 \leq 0. \tag{52}$$

From Eqn 51, Eqn 52, Eqn 48 and Eqn 49, we have that the bounds of mean and variance values of $M_{heu}$ are lower than $M_{ran}$. Given the robust fitness function in Eqn 57, we conclude that

$$Fit_{heu} - Fit_{ran} \leq 0 \tag{53}$$

which is independent of $\varepsilon$ and $\sigma$. In other words, our grouping heuristic can provide tighter fitness bound than random grouping.

## 4. Robust Fitness Function

The makespan (start time of the dummy "sink" activity) for the RCPSP/max with durational uncertainty is a function of non-adjustable variables $\mathbf{x}$ and random variables representing durational uncertainty $\tilde{\mathbf{z}}$ and is represented using $\tilde{S}(\mathbf{x}, \tilde{\mathbf{z}})$ for SLA and $\tilde{G}(\mathbf{x}, \tilde{\mathbf{z}})$ for GNLA. Recall that the robust optimization problem is to find the minimum value $F^*$ for which the following probability bound is observed[5]:

$$P(\tilde{S}(\mathbf{x}, \tilde{\mathbf{z}}) \leq F^*) \quad \geq (1 - \epsilon) \tag{54}$$

From the one-sided Chebyshev's Inequality, we can obtain a bound for the robust objective value $F^*$ as a function of its expected value and variance of the adjustable fitness function, i.e.:

$$E[\tilde{S}(\mathbf{x}, \tilde{\mathbf{z}})] + \sqrt{\tfrac{1-\epsilon}{\epsilon}}\sqrt{Var[\tilde{S}(\mathbf{x}, \tilde{\mathbf{z}})]} \leq F^* \Rightarrow P(\tilde{S}(\mathbf{x}, \tilde{\mathbf{z}}) \leq F^*) \geq (1 - \epsilon) \tag{55}$$

Hence, we can reformulate our robust optimization problem as follows:

$$\begin{aligned} \min \quad & F^* \\ s.t. \quad & E[\tilde{S}(\mathbf{x}, \tilde{\mathbf{z}})] + \sqrt{\tfrac{1-\epsilon}{\epsilon}}\sqrt{Var[\tilde{S}(\mathbf{x}, \tilde{\mathbf{z}})]} \leq F^* \end{aligned} \tag{56}$$

From this model, we can now derive the robust fitness function which will be used in our local search framework:

**Definition 3.** *Given $0 < \epsilon \leq 1$ and the adjustable fitness function $\tilde{S}(\mathbf{x}, \tilde{\mathbf{z}})$ defined above, the robust fitness function, $f(x, \tilde{z}, \epsilon)$, is defined as*

$$f(\mathbf{x}, \tilde{\mathbf{z}}, \epsilon) = E[\tilde{S}(\mathbf{x}, \tilde{\mathbf{z}})] + \sqrt{\frac{1 - \epsilon}{\epsilon}}\sqrt{Var[\tilde{S}(\mathbf{x}, \tilde{\mathbf{z}})]} \tag{57}$$

The goal of the local search mechanism is to find a local minima of $f$. In addition, local search typically requires the fitness function to be computed many times and hence it is imperative that the computation of fitness function is efficient.

---

5. We show the computation of SLA robust fitness function. By substituting $\tilde{S}$ with $\tilde{G}$, we obtain the fitness function for GNLA.

## 4.1 Schedule Infeasibility of a Given POS

It should be noted that the fitness function, $f$ assumes that any schedule generated by the POS, $\mathbf{x}$ is always executable. However, due to durational uncertainty and the maximum time lags, the schedule is not always executable. A direct way to measure $IPr(\text{POS})$ the probability of infeasibility of the POS (i.e. probability that the POS can lead to an infeasible schedule) lies in the computation of the probability of infeasibility of each activity $a_i$ $IPr(a_i)$, that there does not exist a feasible start time such that all temporal constraints with respect to $a_i$ are satisfied. $IPr(\text{POS})$ can be calculated as the probability that at least one activity is infeasible. However, due to temporal dependencies between activities providing a theoretical expression for the overall probability of infeasibility is an open problem. Therefore, we propose a simulation approach, where we simulate POS execution over multiple trials to compute this probability eciently and approximately. As an illustration, we experimented with the benchmark J10 instances from the PSPLib (Kolisch, Schwindt, & Sprecher, 1998) for RCPSP/max with additional durational uncertainty that follows a normal distribution with mean 0 and variance 1. We generated 1000 sample realizations for the POS obtained from SLA, and check for infeasibility with respect to the original temporal (including the maximum time lag) constraints. Examples of the probability of infeasibility obtained by our simulation for PSP1, PSP4, and PSP13 are 0.18, 0.17 and 0.001. However, for the other problems PSP3, PSP5 etc. the probability of infeasibility was 0, because the maximal time lags were much larger than the variance of durational uncertainty.

## 5. Robust Local Search Algorithm

This section will present how the decision rule approximations introduced by SLA, GNLA are integrated with the robust fitness function and local search mechanisms to provide a solution for the problems represented by RCPSP/max with durational uncertainty. Our proposed algorithm is outlined as follows. Steps 1, 2, 5 and 6 are standard steps in a local search algorithm. Steps 3 and 4 represent our departure from standard local search to deal with uncertainty.

1. **Generate initial solution**
   This is usually obtained using a simple greedy heuristic.

2. **Generate neighborhood of solutions**
   Generate a pool of neighbor solutions from the current solution.

3. **Employ one of the decision rule approximations (SLA and GNLA) for all adjustable variables and check feasibility**
   For each candidate solution $\mathbf{x}$ in the solution pool, derive the coefficients $C_k(\mathbf{x})$ for each adjustable variable. Subsequently, for each solution check constraint violation and reject those that are not feasible.

4. **Evaluate robust fitness function $f$**
   For each feasible solution $\mathbf{x}$, evaluate $f$ to obtain the robust objective values. The solution with the lowest robust objective value is the current best robust solution.

5. **Apply penalty (optional)**

   Some advanced local search strategies may require a penalty to be applied to prevent it from being caught at a local minima. In the case of tabu-search for example, a tabu-list is updated when a tabu move is applied. In the case of iterated local search, a perturbation move will be applied to the current local minima.

6. **Termination criteria**

   If the termination criteria is met, return the solution with the lowest robust fitness function value else repeat the optimization cycle by determining the next move.

Algorithm 1 provides the robust local search algorithm guided by decision rule using SLA. By substituting $S^*_{now}, S^*_{min}, S^*$ with $G^*_{now}, G^*_{min}, G^*$, we obtain the local search algorithm using GNLA. Given the RCPSP/max with durational uncertainty and the level of risk ($0 < \epsilon \leq 1$), the algorithm returns the POS with the (locally) minimal robust makespan, $S^*$ (or $G^*$ by GNLA). In essence, we perform robust local search on the neighborhood set of activity lists. An activity list (al) is defined as a precedence-constraint feasible sequence that is used by heuristics to generate earliest start time schedules in solving the standard RCPSP problem (Kolisch & Hartmann, 2005).

Different activity lists are explored by local moves. In our context, we only consider the activity list as the sequence of activities which satisfy the non-negative minimal time lag constraint. Due to the existence of maximal time lag constraint in RCPSP/max, scheduling activities to their earliest possible start time based on the order position in the activity list may restrict the schedule so much that it may not even return in a feasible schedule. Thus, when we schedule each activity sequentially based on order position in the activity list, we will assign its starting time by randomly picking a time from its domain of feasible start times.

According to our experiments, this new randomized approach returns more feasible solutions than the earliest start time one. After finding a feasible schedule, a POS will be generated by applying the chaining procedure proposed by Policella et al. (2004). Then, the $S^*$ (or $G^*$ by GNLA) value will be computed according to the POS. Intuitively, using the randomized approach may return a schedule with a large baseline scheduled completion time. However, we can apply the shortest path algorithm on the resulting POS to generate the earliest start time schedule for a smaller makespan.

As mentioned above, it may be difficult to find a feasible schedule that satisfies minimal and maximal time lag constraints using the activity list. In fact, we believe that in the set of all activity lists, many may not yield a feasible schedule. We overcome this problem as follows. We define the set of activity lists which result in feasible (or infeasible) schedules as $F$ (or $I$). We seek to design a local search algorithm with the following characteristics: a) Starting from an activity list in $I$, the local search should move to an activity list in $F$ within a short time. b) Starting from an activity list in $F$, the local search should move to the activity list with the minimal $S^*$(or $G^*$ by GNLA)value. c) We also diversify the exploration of activity lists in $F$ by allowing the local search to move from an activity list in $F$ to an activity list in $I$, since activity lists in the $F$ region may not be reachable from one another by simple local moves. This has the flavor of strategic oscillation proposed in meta-heuristics research.

---

**Algorithm 1** Robust Local Search

---

1: Generate an activity list **al** randomly
2: Find a start time schedule, $ss$ randomly according to **al**
3: **if al** $\in F$ **then**
4:    $POS \leftarrow chaining(ss)$
5:    Compute $S^*_{now}$ according $POS$
6:    Update $S^*_{min}$ as $S^*_{now}$
7: **else**
8:    Record the first activity $a$ which cannot be scheduled
9: **end if**
10: **for** $i \leftarrow 1$ to Max_Iteration **do**
11:    **if al** $\in I$ **then**
12:       Shift activity $a$ ahead in **al** randomly as **al'**
13:    **else**
14:       Select two activities $b$ and $c$ in **al** randomly
15:       Swap $b$ and $c$ in **al** as **al'**
16:    **end if**
17:    Find randomized start time schedule $ss'$ according to **al'**
18:    **if al'** $\in F$ **then**
19:       $POS' \leftarrow chaining(ss')$
20:       Compute $S^*$ according to $POS'$
21:       **if al** $\in I$ or $S^* \leq S^*_{now}$ **then**
22:          $S^*_{now} \leftarrow S^*$
23:          **al** $\leftarrow$ **al'**
24:          **if** $S^* \leq S^*_{min}$ **then**
25:             $S^*_{min} \leftarrow S^*$
26:          **end if**
27:       **end if**
28:    **else if al** $\in I$ **then**
29:       **al** $\leftarrow$ **al'**
30:    **else**
31:       $p \leftarrow rand(0,1)$
32:       **if** $p < 0.01$ **then**
33:          **al** $\leftarrow$ **al'**
34:          Record the first activity $a$ which cannot be scheduled
35:       **end if**
36:    **end if**
37: **end for**

---

The detailed robust local search procedure is given in Algorithm 1. The procedure starts by randomly generating an activity list **al**, which is a sequence of activities that satisfy the non-negative minimum time lag constraint (Line 1). In Line 2, a schedule *ss* is produced based on ordering of activities in the activity list **al**. We first perform domain reduction on the distance graph using the Floyd-Warshall algorithm, so that the feasible range of the start time for each activity based on the temporal constraints can be obtained. We then schedule each activity sequentially based on the order position in the activity list. For each activity, we first pick a start time randomly from the feasible domain and evaluate resource constraints for the duration of the activity (i.e. check if the current resource capacity exceeds the resource amount used by that activity). If yes, we set the start time to that activity, run the shortest path algorithm to reduce domains for the remaining activities, and update current resource capacity due to consumption of that activity. If the resource constraints are not satisfied, we will try to set the start time randomly again for a prescribed maximum numbers of retries. Once the start time of current activity is set, we proceed iteratively to the next activity according to the activity list. In Line 4, *chaining*() is employed to generate a POS from a baseline schedule (section 2.2.4). *Max_Iteration* refers to the maximum number of iterations in the robust local search. We apply two different types of local moves. To converge quickly to an activity list in F, the first local move is designed to schedule the activity that is causing a temporal or resource conflict to an earlier time. It will randomly shift ahead the first activity which cannot be scheduled in the current activity list (Line 12). When an activity list is in F, the second local move will randomly pick two activities and swap them in the current activity list, while satisfying the non-negative minimal time lag constraints (Line 14-15). The move will be accepted, if it results in a smaller or equal $S^*$ value (Line 18-29). To explore different activity lists, we include a small probability to accept the move which leads to an infeasible schedule (Line 31-35). The probability to move from an activity list in F to one in I is set at 0.01. The minimal $S^*$ value will be saved as $S^*_{min}$.

The worst-case computational complexity analysis is given as follows. For each iteration in local search, there are three major components: randomized schedule generation, POS construction and fitness calculation. In the process of randomized schedule generation, we perform domain reduction and resource checking at each iteration, and thus the complexity is $O(N \cdot (N^3 + H \cdot K \cdot w))$ where $N$ is the number of activities, $H$ is the maximum planning horizon, $K$ is the number of types of resources, and $w$ is the prescribed maximum number of retries for each activity on setting the randomized start time. The POS construction process works as follows: the set of activities are first sorted according to their start times in the generated deterministic schedule and the sorting part costs $O(N \cdot log N)$; then it proceeds to allocate each activity the total units needed for each type of resource. Let $max_{cap}$ be the maximum capacity among all resources. The cost for computing POS is then $O(N \cdot log N + N \cdot K \cdot max_{cap})$. When determining the fitness value of generated POS, we examine edge by edge to check if it is connected in parallel or in serial with respect to its predecessors and it costs $O(N + e)$ where $e$ is the number of edges in POS ($e < N^2$). Thus, the worst-case complexity of our proposed robust local search algorithm is $O(TN \cdot (N^3 + H \cdot K \cdot w + K \cdot max_{cap}))$ where $T$ is the number of iterations in local search.

## 6. Enhancing Robust Local Search

In this section, we describe two enhancements to improve the basic local search method described in Section 5. Firstly, we describe ordering generation, which is a pre-processing step used to identify precedence ordering between activities. This precedence ordering is then used to focus the local search over activity lists. Secondly, we describe a new chaining method to generate POS from a feasible schedule.

### 6.1 Ordering Generation

Ordering Generation is a pre-processing step that identifies precedence relationships between pairs of activities. The key idea is that for certain pairs of activities, it is always better (with respect to robust makespan) to have the same ordering among activities. Our goal is to identify these pairs of activities and employ this ordering to focus the local search over activity lists and in the chaining method used to compute POS from feasible schedule.

In deciding an ordering between a pair of activities, $a$ and $b$, there are two key steps: (i) Sample set generation: Generate two sets of $m$ activity lists. The first set consists of $m$ activity lists where $a$ occurs before $b$. The second set is generated by swapping activities $a$ and $b$ in every activity list in the first set; (ii) Order determination: In this step, we first compute POS and its robust makespan for all activity lists in the two sets. By comparing the robust makespan values of corresponding activity lists in the two sets, we determine an ordering between activities. We explain these steps in the following subsections.

For a problem with $n$ activities, there are $C_n^2$ pairs of activities. If we are to decide the orders between all pairs, the ordering computation needs to be implemented for $C_n^2$ times, which is computationally expensive. Based on this observation, we first propose a *Pairs-Selection* heuristic to selectively choose a certain number of activities pairs whose ordering can have a significant impact on the robust makespan.

The *Pairs-Selection* heuristic picks an activity pair: (a) If it is not precedence related in the original problem definition; and (b) If there exists at least one type of resource, where the total demand of both activities exceeds the resource capacity. The intuition behind picking such an activity pair is that those two activities cannot be executed in parallel and deciding an ordering relationship is imperative to eliminate the resource conflict. One main advantage of the heuristic is that the number of pairs of activities that need to be ordered is significantly reduced. Now, we describe the two steps of ordering generation below:

### 6.1.1 Sample Set Generation

We first randomly generate $m$ activity lists as an initial sample set denoted by $T$. Each element in $T$ is an activity list represented as $al_i$ which is a sequence of all activities, where $i = 1, ...m$, i.e.

$$T = \{al_i | al_i = (a_1, a_2, ...a_n), \forall i \in \{1, ...m\}\}.$$

For each pair of activities $(a_k, a_l)$ resulting from the *Pairs-Selection* heuristic, we define two sample sets represented as $T^{a_k \prec a_l}$ and $T^{a_l \prec a_k}$. $T^{a_k \prec a_l}$ has all the activity lists that are in $T$, except that if an activity list has $a_l$ before $a_k$, then those activities are swapped.

$$T^{a_k \prec a_l} = \{al_i^{a_k \prec a_l} | i \in \{1, ...m\}\},$$

where $al_i^{a_k \prec a_l} = \begin{cases} (a_1, a_2, ..., a_k, ...a_l, ...a_n) & if \ \ al_i = (a_1, a_2, ..., a_l, ...a_k, ...a_n) \\ al_i & if \ \ al_i = (a_1, a_2, ..., a_k, ...a_l, ...a_n) \end{cases}$.

Similarly, $T^{a_l \prec a_k}$ can be constructed by incrementally selecting each activity list from the initial set $T$ with $a_l \prec a_k$ and reverse the order if $a_k \prec a_l$, i.e.

$$T^{a_l \prec a_k} = \{al_i^{a_l \prec a_k} | i \in \{1, ...m\}\},$$

where $al_i^{a_l \prec a_k} = \begin{cases} (a_1, a_2, ..., a_l, ...a_k, ...a_n) & if \ \ al_i = (a_1, a_2, ..., a_k, ...a_l, ...a_n) \\ al_i & if \ \ al_i = (a_1, a_2, ..., a_l, ...a_k, ...a_n) \end{cases}$.

Thus, each activity list in the sample set $T^{a_k \prec a_l}$ share the same positions of all activities except $a_k$ and $a_l$ with the corresponding activity list in set $T^{a_l \prec a_k}$, where $a_l$ precedes $a_k$.

### 6.1.2 Order Determination

We then determine the activity order of each selected pair of activities based on the sample sets obtained from last phase. For a pair $(a_k, a_l)$, we construct a new instance by posting a precedence constraint $a_k \prec a_l$ or $a_l \prec a_k$ to the original instance, and based on the new instance, we determine the fitness which are denoted as $f_i^{a_k \prec a_l}$ and $f_i^{a_l \prec a_k}$ for $al_i^{a_k \prec a_l}$ and $al_i^{a_l \prec a_k}$, respectively.

Note that $al_i^{a_k \prec a_l}$ and $al_i^{a_l \prec a_k}$ share the same elements and the same positions except the order of $a_k$ and $a_l$. Thus, the order of $a_k$ and $a_l$ can be considered as a reason why the fitness of $al_i^{a_k \prec a_l}$ and $al_i^{a_l \prec a_k}$ differs. To decide the order of $a_k$ and $a_l$, we define an *index variable* denoted as $iv_{a_k \prec a_l}$ that measures the percentage of samples where the one with the order $a_k$ proceeds $a_l$ wins, i.e.

$$iv_{a_k \prec a_l} = -\frac{\sum_i min(\frac{f_i^{a_k \prec a_l} - f_i^{a_l \prec a_k}}{|f_i^{a_k \prec a_l} - f_i^{a_l \prec a_k}|}, 0)}{m}.$$

We then define an *Index Parameter* for activities $a_k$ and $a_l$ denoted as $IP_{a_k \prec a_l}$ as a benchmark for the index variable $iv_{a_k \prec a_l}$ in determining the order of $a_k$ and $a_l$. The parameter $IP_{a_k \prec a_l}$ can be prescribed by users and different values (usually larger than 50%) represent different levels of confidence that the order of $a_k$ and $a_l$ matters in causing fitness variance, and thus also represents different controllability of $iv_{a_k \prec a_l}$.

If the value of index variable $iv_{a_k \prec a_l}$ is larger than the value of $IP_{a_k \prec a_l}$, we set the order $a_k \rightarrow a_l$ since there indicates a higher probability that $a \rightarrow b$ can provide better robustness than $b \rightarrow a$; If $iv_{a_k \prec a_l}$ is less than $1 - IP_{a_k \prec a_l}$, we set $a_l \rightarrow a_k$; And in other cases, no order between $a_k$ and $a_l$ is settled.

## 6.2 Improved Chaining based on Robustness Feedback

As noted in the "Preliminaries" section, for each activity $a$, there may exist multiple choices of resource chains to which it can be assigned. In addition, different chaining heuristics will lead to POSes that can have different robust makespan values. In this section, we propose a new chaining heuristic that dispatches activities to resource chains by predicting the improvement in robust makespan of the generated POS.

---

**Algorithm 2** Robustness-Feedback Resource Chaining (Activity $a$, Schedule $S$, Order $G$)

---
1: $C \leftarrow$ Find set of available chains, C for activity $a$ based on $S$
2: $P \leftarrow$ Collect chains from C with last activity of chain preceding $a$ in problem
3: $O \leftarrow$ Collect chains from C with last activity of chain ordered before $a$ in $G$
4: **if** $P \neq \phi$ **then**
5:     $k \leftarrow$ Get first available chain in $P$
6: **else if** $O \neq \phi$ **then**
7:     $k \leftarrow$ Get first available chain in $O$
8: **else**
9:     $k \leftarrow$ Get first available chain in $C$
10: **end if**
11: Post constraint between last activity of chain $k$ (denoted as last(k)) and activity $a$
12: **if** $a$ requires more than one resource unit **then**
13:     $C1 \leftarrow$ chains in $C$ which have last activity as last(k)
14:     $C2 \leftarrow C \setminus C1$
15:     **for all** resource units required by $a$ **do**
16:        choose the first available chain belonging to $C1$
17:        **if** chain above is not feasible **then**
18:           choose the first available chain belonging to $C2$
19:        **end if**
20:     **end for**
21: **end if**

---

In the latest chaining method which aims to increase flexibility as described in Section 2.2.4, the chains are first randomly picked from a superior subset (i.e., chains where the last activity is already ordered, or chains sharing the same last element). Since our objective is makespan-related and time becomes a concern, we build on the work of Policella et al. (2009) and pick the *first* available chain wherever available. The updated chaining method is called *Robustness-Feedback based Resource Chaining*.

**Example 5.** *Figure 5 provides the POS provided by this chaining heuristic when used on Example 1. As can be seen, compared to the POS in 4, the key difference is the allocation of activity 5 and 6. With our new heuristic, it can be seen that there is more parallelism and hence reduced robust makespan with high probability.*

When employing the Ordering Generation algorithm in conjunction with the chaining heuristic, we also consider the information about ordered pairs when allocating resource units to an activity. The motivation is that once activity $a$ and activity $b$ (for example, $a \rightarrow b$) is ordered, there is a high probability that this precedence relationship can result in a better solution. Algorithm 2 provides the pseudo code for the Robustness-Feedback Resource Chaining heuristic with Ordering.

## 7. Experimental Evaluation

In this section, we first evaluate the scalability and quality of the execution strategies provided by robust local search and the various enhancements introduced in this paper.

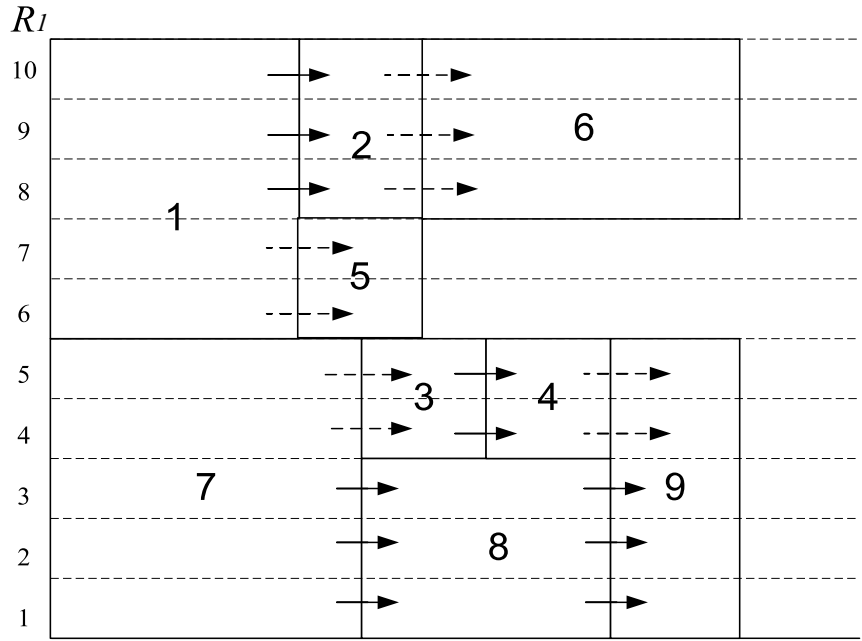

Figure 5: POS by Robustness Feedback Chaining

Secondly, to establish a benchmark on the performance, we compare against the best known technique for solving JSP problems with durational uncertainty. It should be noted that the robust local search method is developed to solve RCPSP/max problems with durational uncertainty and hence does not exploit the structure present in JSP problems. Furthermore, as described earlier, the optimization metrics of both approaches are different.

## 7.1 Experimental Setup

We have two sets of problems that we consider and those are described in the subsections below. Additionally, we also indicate the algorithms that are compared on each of the data sets in this section.

### 7.1.1 RCPSP/max with Durational Uncertainty

The problems considered for RCPSP/max with durational uncertainty were obtained by extending the three benchmark sets available for RCPSP/max problems, J10, J20 and J30 as specified in the PSPLib (Kolisch et al., 1998). Each set contains 270 problem instances with duration for each activity ranging between 1 and 10. The maximum number of activities for J10, J20 and J30 are 10, 20 and 30, respectively. For each activity $a_i$, we set the expected value $d_i^0$ of the stochastic duration as the corresponding deterministic duration given by the benchmarks, and assume that duration uncertainty is normally distributed, i.e. $\tilde{z}_i \sim N(0, \sigma)$. Henceforth, we refer to J10, J20 and J30 as these RCPSP/max problems with durational uncertainty. We run the algorithms on problems with four different duration variabilities $\sigma = \{0.1, 0.5, 1, 2\}$ and four increasing levels of risk $\varepsilon = \{0.01, 0.05, 0.1, 0.2\}$.

On RCPSP/max problems with durational uncertainty, we compare the robust local search that is guided using the two decision rule approximations SLA and GNLA. Furthermore, we also compare the different enhancements to robust local search on RCPSP/max problems with durational uncertainty. We compare five different variants of robust local search for each decision rule approximation: (a) (GNLA) refers to basic robust local search guided by GNLA decision rule approximation; (b) (GNLA+RC) is the robust local search with the new Robustness-feedback Chaining heuristic guided by GNLA; (c) (GNLA+) refers to the basic robust local search with additional local search iterations, where the number of local search iterations is determined based on the problem set (as described later); (d) (GNLA+OG) is the Order Generation heuristic on top of GNLA guided robust local search; and finally (e) (GNLA+OG+RC) has both Order Generation and Robustness-feedback Chaining heuristics on GNLA guided robust local search.

The number of local search iterations for robust local search was set to 1000. To reduce the stochasticity effects of robust local search, we average over 10 random executions for each problem instance. Our code was implemented in C++ and executed on a Core(TM)2 Duo CPU 2.33GHz processor under FedoraCore 11 (Kernel Linux 2.6.29.4-167.fc11.i586).

### 7.1.2 JSP with Durational Uncertainty

For JSPs, (GNLA) is compared against the probabilistic makespan results provided by Beck and Wilson (2007). For the benchmark problems, we consider the instances generated using an existing generator in the work of Watson, Barbulescu, Whitley, and Howe (2002) with durations drawn uniformly from the interval [1,99]. Specifically, we focus on three sets of probabilistic JSPs of size $\{4\times4,6\times6,10\times10\}$ (where a $4\times4$ problems consists of 4 jobs consisting of 4 activities each) and for each set, three uncertainty levels $\{0.1,0.5,1\}$ are considered.

## 7.2 Comparison between SLA and GNLA

We first compare the average robust makespan of 270 problem instances obtained by robust local search that is guided by our decision rule approximations proposed in Section 3.1 and Section 3.2. We refer to the robust makespan computed using SLA as $S^*$ and using GNLA as $G^*$. Figure 6 provides these results for all three sets of RCPSP/max problems with durational uncertainty. In these results, we show how the robust makespan is affected by the level of risk $\varepsilon$ and the standard deviation $\sigma$ of duration uncertainty. X-axis represents different combinations of risk and standard deviation of durational uncertainty, as shown in the table of Figure 6. All runs on every instance takes a couple of seconds and hence we do not report CPU times here. The key observations and conclusions of interest from Figure 6 are as follows:

- Irrespective of the $\sigma$, as the level of risk $\varepsilon$ increases, the robust makespan decreases with both SLA and GNLA. Clearly, the lower risk that the planner is willing to take, the higher is the robust value of the generated execution strategy. Our method is capable of quantifying this trade off, which can help the planner to decide on the desired strategies.

| Horizontal Scale | 1 | 2 | 3 | 4 | 5 | 6 | 7 | 8 | 9 | 10 | 11 | 12 | 13 | 14 | 15 | 16 |
|---|---|---|---|---|---|---|---|---|---|---|---|---|---|---|---|---|
| Sigma x Epsilon | 0.1 x 0.01 | 0.5 x 0.01 | 1 x 0.01 | 2 x 0.01 | 0.1 x 0.05 | 0.5 x 0.05 | 1 x 0.05 | 2 x 0.05 | 0.1 x 0.1 | 0.5 x 0.1 | 1 x 0.1 | 2 x 0.1 | 0.1 x 0.2 | 0.5 x 0.2 | 1 x 0.2 | 2 x 0.2 |

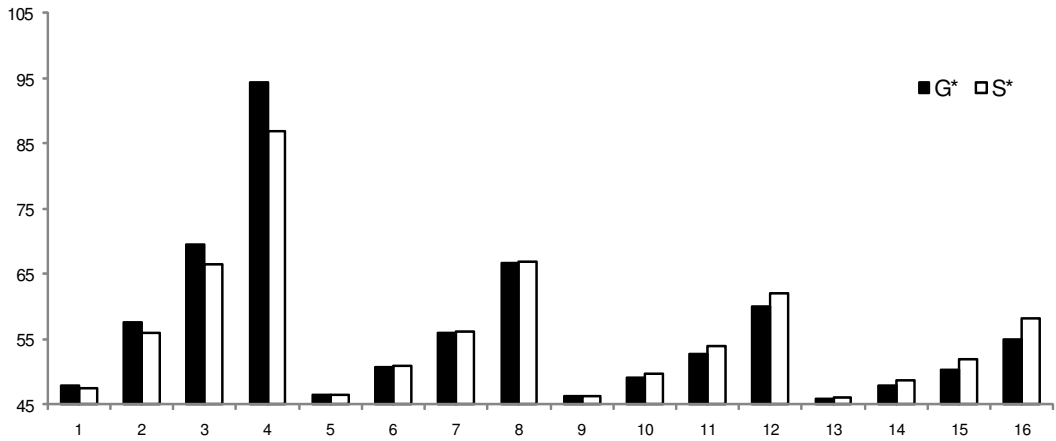

(a) Results of J10

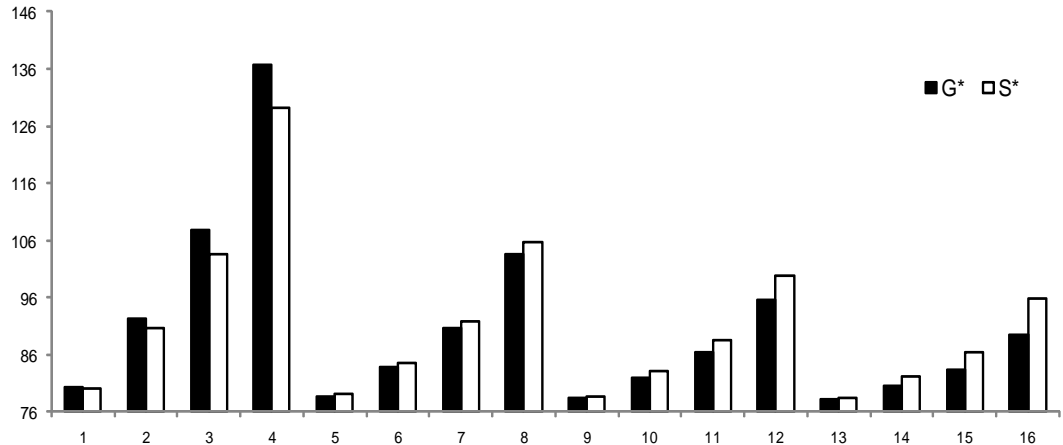

(b) Results of J20

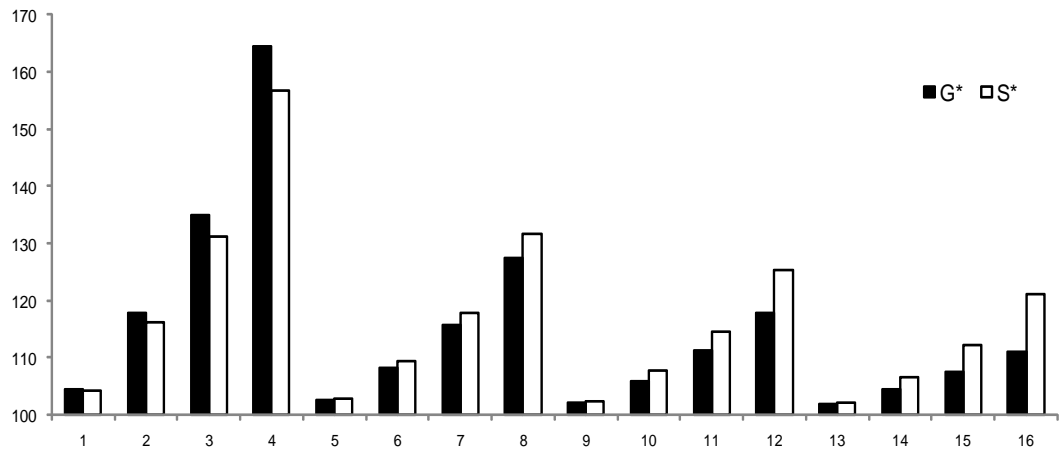

(c) Results of J30

Figure 6: Comparison of Robustness between SLA and GNLA.

- Irrespective of $\varepsilon$, as the degree of duration variability $\sigma$ increases, the robust makespan increases with both SLA and GNLA, and the value becomes more sensitive to $\sigma$ when the level of risk is constrained to a small value (e.g. $\varepsilon = 0.01$).

- For lower values of $\varepsilon$, more specifically for $0.01$, $S^*$ provides lower values of robust makespan than $G^*$. On the other hand, for higher values of $\varepsilon \in \{0.05, 0.1, 0.2\}$, $G^*$ provides superior performance to $S^*$. We do not yet understand the reason for drop in performance for $\varepsilon = 0.01$, but this is observed consistently across all the RCPSP/max benchmark problems.

For each problem instance, we also observe some monotonicity between the absolute difference of robust makespan $S^*$ and $G^*$ and risk values. When the level of risk $\varepsilon$ takes a value around $0.02$, $S^*$ (SLA) has a slightly lower value than $G^*$ (GNLA). However, when risk becomes more than $0.02$, the superiority of GNLA increases with higher values of risk. Figure 7 illustrates this on a randomly picked J10 instance with $\sigma = 1$ and $\sigma = 2$. The same pattern is observed across all problem instances of J10, J20 and J30.

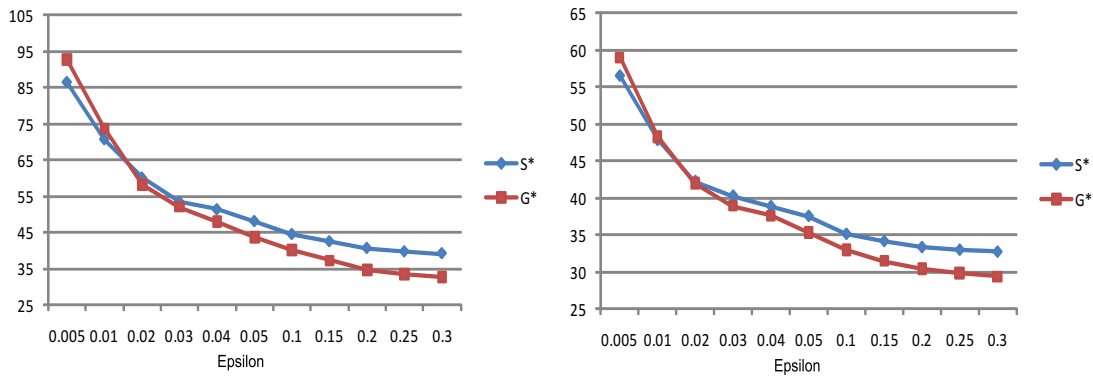

(a) Results from a randomly selected J10 example with $\sigma$=1

(b) Results from a randomly selected J10 example with $\sigma$=2

Figure 7: Comparison of Robust Makespan.

Next, in Figure 8, we compare the quality of the execution strategies obtained by using SLA and GNLA. More precisely, we compare the distributions of the actual makespans of schedules computed using these decision rule approximations. For this purpose, we generate a set of 100 samples of realizations of durational uncertainty and test with all 270 instances of each benchmark set with different levels of risk $\varepsilon = 0.2$, $\varepsilon = 0.1$ and $\varepsilon = 0.05$ to obtain the respective POS, and then compute the actual makespans of schedules derived from the respective POS under the given realization samples. This difference between real makespans obtained from POSs generated by two different decision rule approximations was observed across the board in all examples of the three sets for all values of $\varepsilon$ except $0.01$. We randomly select three problem instances from each benchmark set and present the results in Figure 8. Figure 8 also compares the cumulative frequency distributions of the actual makespans. We observe that GNLA provided far better realized makespans than SLA - both in absolute terms, as well as distributionally. For J20, except in 2 cases, rest of the actual makespan

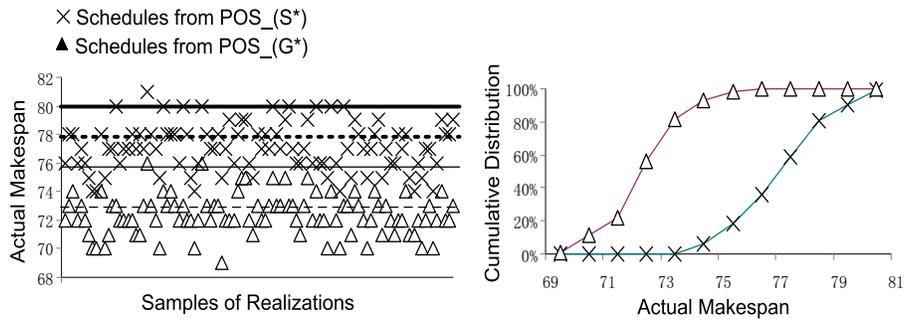

(a) Results from randomly selected J10 example with $\varepsilon = 0.2$

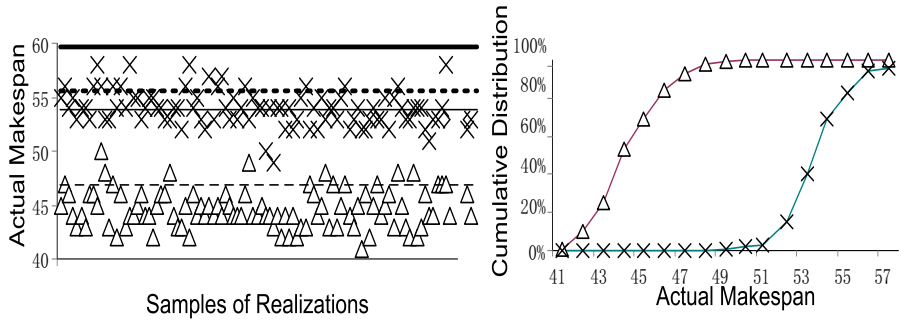

(b) Results from a randomly selected J20 example with $\varepsilon = 0.1$

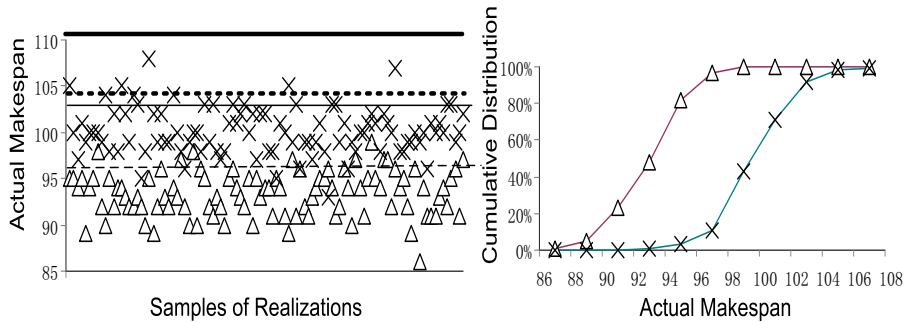

(c) Results from a randomly selected J30 example with $\varepsilon = 0.05$

Figure 8: Comparison of Actual Makespans and Gap between $S^*$ and $G^*$.(Lines in left pictures from top down indicating: Computed $S^*$, Actual $S^*$ by Simulation, Computed $G^*$, Actual $G^*$ by Simulation.)

values obtained by SLA were higher than the ones obtained by GNLA. Similar trends were observed for J10 and J30.

To illustrate the difference of quality in absolute of the two upper bounds, we provide four lines (computed $S^*$, actual $S^*$, computed $G^*$ and actual $G^*$) indicating the upper bounds computed using the algorithms and in simulation over the 100 samples.

### 7.3 Comparing Robust Local Search Enhancements

Since, we have already shown GNLA performs better than SLA, we will only show the performance of our enhancements over GNLA in this section. It should be noted that enhancements over SLA provided similar results and conclusions with GNLA based enhancements outperforming SLA based enhancements. Since "Ordering Generation" heuristic requires additional rounds of robust makespan computation, we also include a benchmark called (GNLA+) (which is GNLA plus extra iterations of local search) to make a fair comparison. To avoid the complexity of considering all pairs of activities, we only consider those pairs of activities where ordering would improve performance. We proposed the Pairs-Selection heuristic to select these pairs of activities. The number of extra iterations of local search for the (GNLA+) benchmark is "the number of activity pairs picked by the Pairs-Selection heuristic" times "the number of samples $m$ used in the Ordering Generation process". The experimental results shows that the average number of activity pairs of all 270 instances selected under the Pairs-Selection heuristic for J10, J20 and J30 are 5, 14, and 28 respectively. In our work, we set $m = 100$. Thus, the extra iterations of the (GNLA+) benchmark for J10, J20 and J30 are 500, 1400 and 2800, respectively. The performance of all our enhancements is shown in Figure 9(a), Figure 9(b), Figure 9(c) for J10, J20 and J30 respectively. In all the charts, $\varepsilon$ is represented on the X-axis and robust makespan on the Y-axis. So, lower values are better on the Y-axis.

Given below are some key observations and conclusions made from the results:

- Irrespective of the durational uncertainty, (GNLA+RC) and (GNLA+OG) provide better robust makespan values than both (GNLA) and (GNLA+) for J10 and J30. This indicates that the new Robustness Feedback Chaining heuristic and the Order Determination are able to provide more robust partial ordered schedules for J10 and J30. This improvement seems to increase further with more number of activities, i.e. the difference is more obvious for instances in J30 than in J10. Furthermore, the difference is consistently observed across all the problems. However, the improvement is not consistent for J20 and there are cases where (GNLA+RC) and (GNLA+OG) did not out perform (GNLA) and (GNLA+). For instance in J20 problems, (GNLA+) provides better performance than (GNLA+RC) and (GNLA+OG) for $\varepsilon = 0.01$ and $\sigma = 1.5$.

- The extra iterations of local search in (GNLA+) do not improve the solution quality much for J10. However, it improves the solution quality for J20 and J30. This could be because the optimal solution is obtained within 1000 iterations for the smaller problems.

- In most cases, (GNLA+RC+OG) provides the lowest robust makespan among all the enhancements. Thus, the OG and RC enhancements in combination do not degrade the performance improvement obtained individually. In some cases, the difference is significant such as in J10 for $\sigma = 0.1$ and $\varepsilon = 0.01$. On the other hand, there are cases where (GNLA+RC+OG) does not provide the lowest robust makespan, such as in J20 for $\sigma = 0.5$ and $\varepsilon = 0.01$.

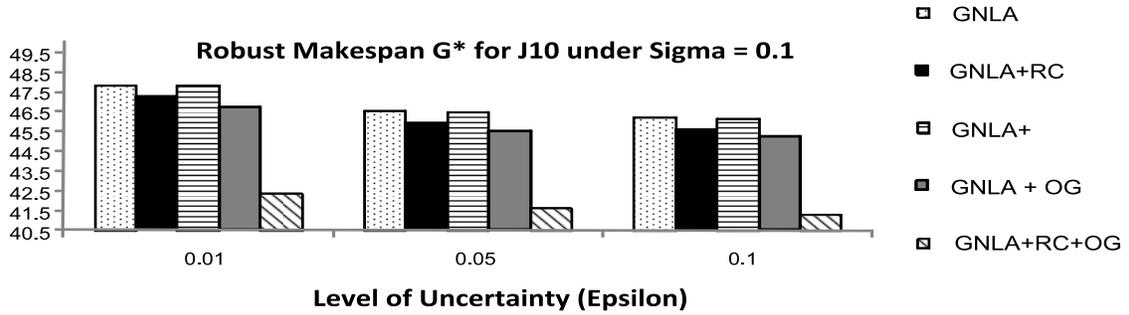

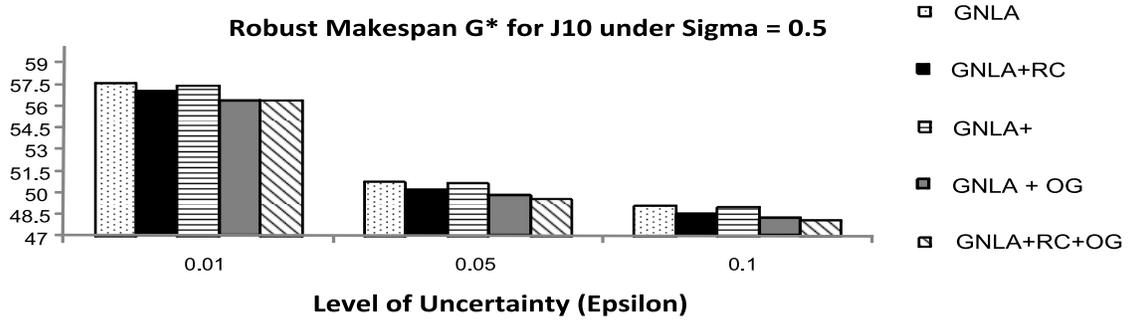

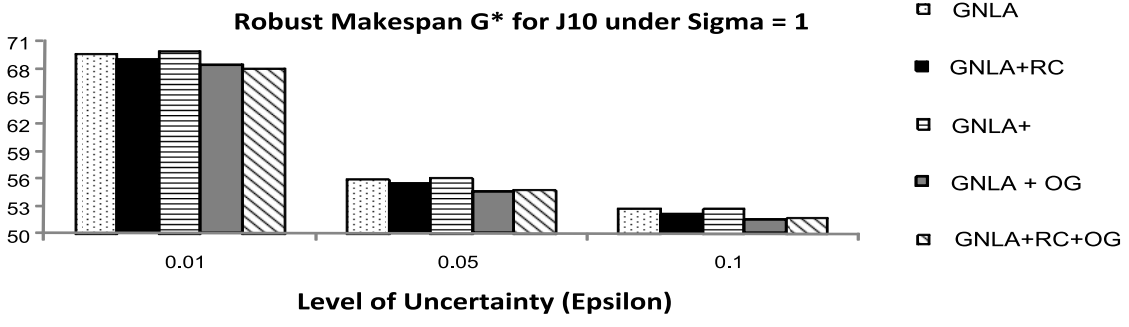

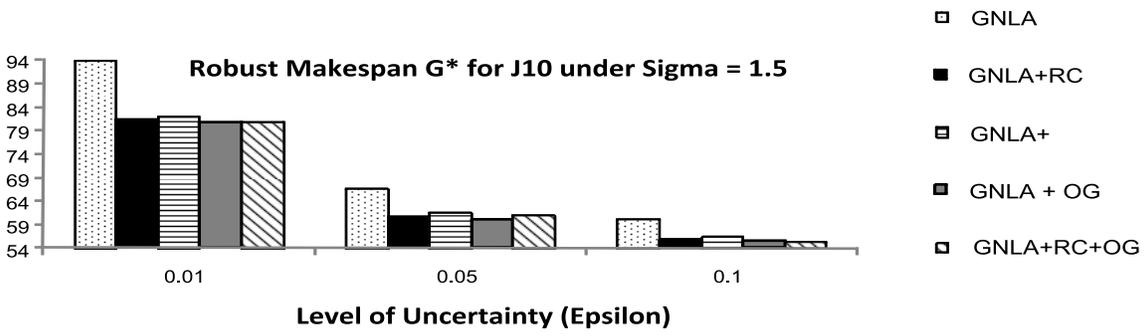

(a) Results of J10

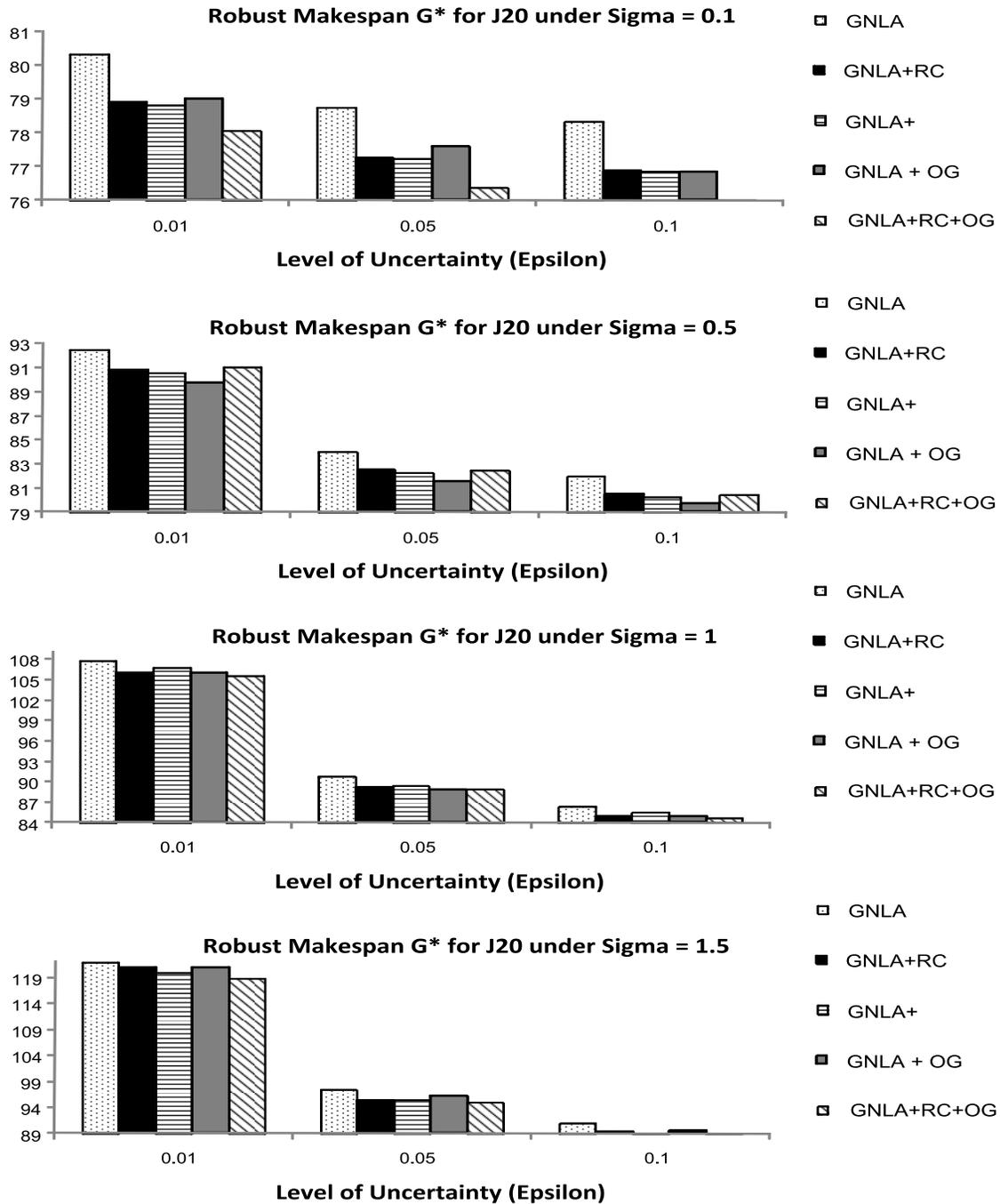

(b) Results of J20

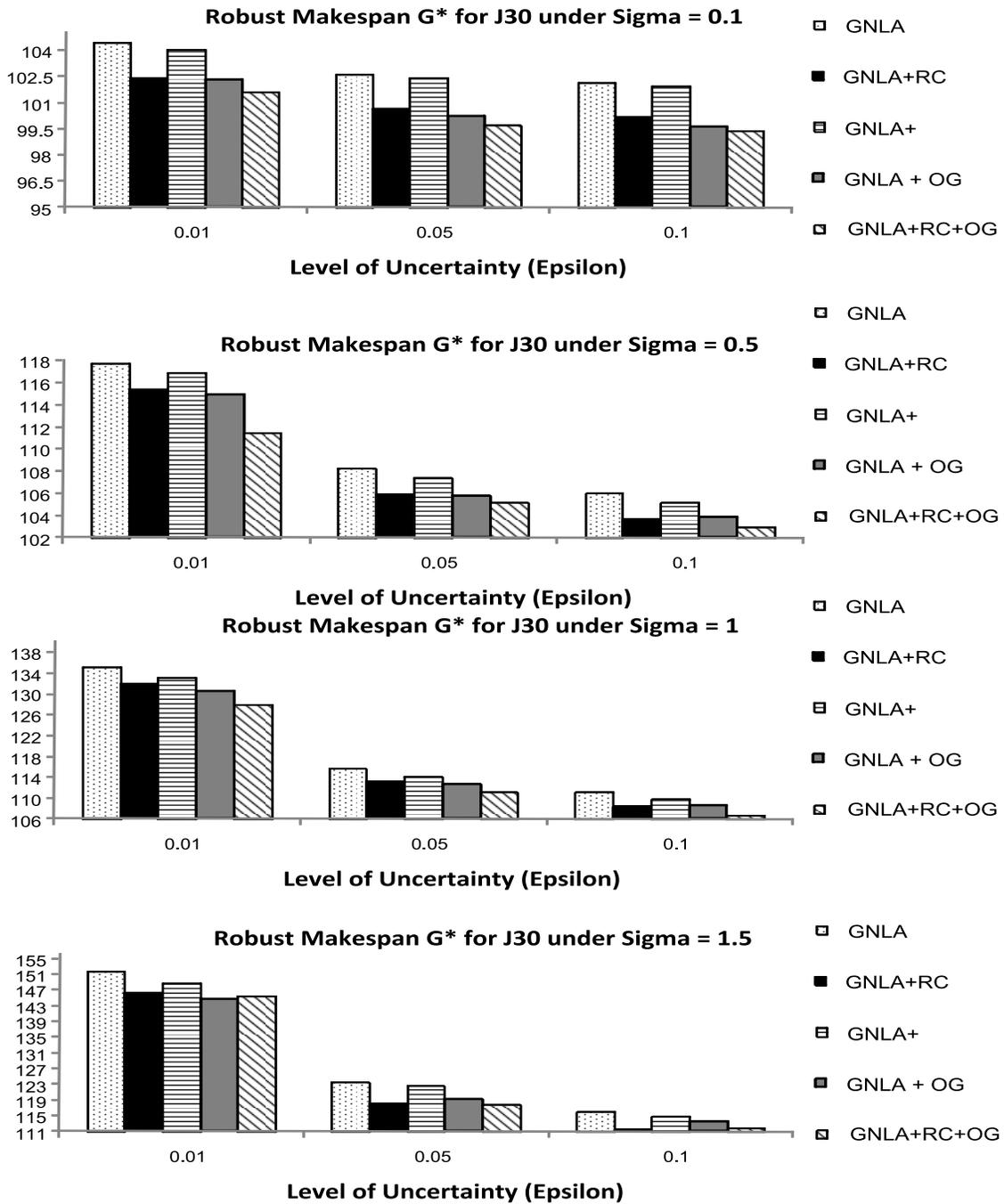

(c) Results of J30

Figure 9: Comparison of Robust Local Search Enhancements.

| MNPM | Problem Size 4×4 | | | Problem Size 6×6 | | | Problem Size 10×10 | | |
|------|--------|--------|------|--------|--------|------|--------|--------|------|
|      | UL=0.1 | UL=0.5 | UL=1 | UL=0.1 | UL=0.5 | UL=1 | UL=0.1 | UL=0.5 | UL=1 |
| CB   | 1.023  | 1.046  | 1.128| 1.021  | 1.073  | 1.168| 1.024  | 1.101  | 1.215|
| $G^*$| 1.066  | 1.123  | 1.282| 1.095  | 1.190  | 1.273| 1.210  | 1.225  | 1.263|

Table 2: Comparison against CB solver(UL:Uncertainty Level).

## 7.4 Comparing on JSPs with Durational Uncertainty

In this section, we compare the performance of our GNLA approach (referred to as $G^*$) with the best known solver for Job Shop Scheduling Problems proposed by Beck and Wilson (2007) (referred to as CB). For a fair comparison of the two approaches, we employ the Mean Normalized Makespan (MNPM) metric defined by Beck and Wilson:

$$MNPM(a, L) = \frac{1}{|L|} \sum_{l \in L} \frac{D(a, l)}{D_{lb}(l)} \qquad (58)$$

where L is a set of problem instances, $D(a, l)$ is the probabilistic makespan (i.e., robust makespan in our work) for instance $l$ by algorithm $a$ generated by Monte Carlo simulation, $D_{lb}(l)$ is a lower bound on the probabilistic makespan.

We denote the best MNPM values aross different algorithms reported by Beck and Wilson as CB. We compare them with the MNPM values in our work which are obtained by replacing $D(a, l)$ in Eqn 58 with an upper bound of robust makespan from the POS generated from GNLA-guided local search. All runs on $4 \times 4$ and $6 \times 6$ instances took less than a minute, while $10 \times 10$ instances took about 15 minutes.

Table 2 provides the results. The performance of our solver is comparable to CB solver across all problem instances. This comparison illustrates that our local search mechanism is generic (different types of scheduling problems) and is also able to provide performance on par with near optimal approaches. While the performance is comparable, CB provides better MNPM values than our approach due to the following key reasons: (a) Our approach does not exploit the structure specific to JSPs (jobs consisting of a sequence of operations). We hope to improve our approach to exploit this in the near future. (b) Our robust local search reasons with upper bounds (due to Chebyshev inequality), which can be loose.

## 8. Related Work

The Resource-Constrained Project Scheduling Problem with minimum and maximum time lags, RCPSP/max, (or known as the Resource-Constrained Project Scheduling Problem with Generalized Precedence Relations, RCPSP-GSR) is a strongly NP-hard combinatorial optimization problem; and even the decision problem of determining whether an RCPSP/max instance has a feasible solution or not is NP-complete (Bartusch et al., 1988). A survey of recent developments and new applications for RCPSP/max has been given by Neumann, Schwindt, and Zimmermann (2006).

However, we did not find much study that considers RCPSP/max under uncertainty. One such paper dealing with variable durations on RCPSP/max is done by Lombardi and Milano (2009), where activity durations range between given lower and upper bounds. A precedence constraint posting approach (Policella, Cesta, Oddi, & Smith, 2007) was adopted. Whereas in our work, we consider RCPSP/max with durational uncertainty where each activity duration is modeled as a random variable with known mean and variance values.

Research on scheduling under uncertainty has received much attention in both Artificial Intelligence and Operations Research communities. For a complete survey of recent AI papers on robust project scheduling up to 2004, one may refer to the work of Herroelen and Leus (2005) and of production scheduling (Aytug, Lawley, McKay, Mohan, & Uzsoy, 2005). Broadly, one may classify the techniques to tackle scheduling with uncertainty into two categories: *Proactive Scheduling* is to design a priori schedule or a schedule policy that take into account the possible uncertainty that may occur; *Reactive Scheduling* modifies or re-optimizes the baseline schedule when an unexpected event occurs. Here our interest is on proactive scheduling and we are concerned with robust scheduling which focuses on obtaining proactive schedules that maintain a high level of performance against uncertainty.

The main idea of *proactive* techniques is to build a global solution which hopefully does not need to be revised at execution time. One can divide the research in this area into three categories, according to how and when the information of uncertainties can be taken into account in generating more robust and stable schedules than they would be without using this information (Bidot, Vidal, Laborie, & Beck, 2009): 1. generating one complete *generic* schedule which is proved to execute correctly in most scenarios arising during execution; 2. generating a *flexible* solution in which some decisions are postponed to be made until execution; 3. generating a *conditional* solution in which mutually exclusive decisions are developed,the one being chosen dependent on some observations during execution, like markov decision processes. In the following, we briefly look at the first two cases since they are related to our work.

## 8.1 Generating Generic Schedule

A first method for making a *generic* schedule that is insensitive to online perturbations is to produce a complete and robust schedule by taking into account all possible scenarios, i.e. a schedule with *strong controllability* (Vidal & Fargier, 1999). Rather than dealing with execution with 100% confidence, *probabilistic techniques* have been proposed that build schedules with a probabilistic guarantee against a threshold value of an optimization metric such as makespan. Another example of such generic schedule generation is *fuzzy scheduling* (Herroelen & Leus, 2005): instead of stochastic variables and probabilistic distributions, fuzzy set scheduling use fuzzy numbers for modeling uncertainties based on possibility theory; a recent work by Rodríguez et al. (2009) modeled uncertain durations as fuzzy numbers and improved local search to solve the Job Shop Scheduling Problem. In the following, we provide further details on the work related to strong controllability and probabilistic techniques.

### 8.1.1 Strong Controllable Techniques

Strong Controllability was introduced by Vidal and Fargier (1999) over Simple Temporal Networks with Uncertainty (STNU) for which controllability is achievable in polynomial time. With the existence of uncontrollable events that are controlled by exogenous factors, often referred to as Nature, an STNU is strongly controllable if there exists at least one universal schedule that suits any situation. Such schedule might be computed off-line beforehand. Strong controllability is the strictest form of STNU. A strongly controllable network means that the schedule can be executed without regard to the contingent events. It is useful in applications where contingent events cannot be observed exactly.

### 8.1.2 Probabilistic Techniques

Instead of generating a global solution suitable for all realizations of uncertainties, probabilistic techniques build a schedule that has a probabilistic guarantee of a deterministic optimization measure with respect to a threshold value, e.g., find the schedule with the highest probability that the project makespan will not exceed a particular value.

Daniels and Carrillo (1997) defined a $\beta$-robust schedule as one that has maximum probability of achieving a given performance level, e.g., the total flow time is no greater than a given threshold. They presented branch-and-bound and heuristic techniques to find a robust schedule in a one-machine manufacturing context that performs the best within a given confidence level. As for the Job Shop Scheduling Problem, Beck and Wilson (2007) consider activity durations as random variables; given a level of risk $0 \leq \alpha \leq 1$, they are interested in a solution of minimal (probabilistic) makespan which has a probability of execution of at least $1 - \alpha$.

## 8.2 Generating Flexible Schedule

Another way of producing robust schedule taking into account of uncertainty is to introduce flexibility into the schedule. The idea is that only a subset of decisions are made offline and the rest are postponed to be made online, so that decisions are only made when information becomes more precise and certain (Bidot et al., 2009). In the following, we discuss three subcategories of works that deal with generating flexible schedules.

### 8.2.1 Dynamic Controllable Techniques

An STNU is Dynamic Controllable (Vidal & Fargier, 1999) if there exists a solution that can always be instantiated incrementally based on the outcomes of contingent edges in the past. An execution strategy using dynamic controllability is needed to produce an incremental solution based on the subsequent revelation of contingent events. Morris and Muscettola (2005) proposed a pseudo-polynomial algorithm to handle dynamic controllability of STNUs based on constraint satisfaction. Techniques were proposed by Wah and Xin (2004) to optimize the bounds on durations of contingent edges such that the resulting STNU is dynamic controllable.

### 8.2.2 Redundancy-based Techniques

Redundancy-based scheduling is another proactive technique for scheduling. The idea is to generate a schedule that includes the allocation of extra resources and/or time in the schedule so that these buffers will help absorb the impact of unexpected events without rescheduling during execution. Davenport, Gefflot, and Beck (2001) proposed techniques for generating robust schedules based on the insertion of temporal slacks to critical activities that are allocated on possibly breakable resources. Lambrechts, Demeulemeester, and Herroelen (2010) analytically determined the expected increase in activity duration due to resource breakdown. Based on this information, simulation-based time buffering was used to protect the schedule from disruptions caused by resource availability.

### 8.2.3 Partial Order Schedule (POS)

Even with buffering, baseline schedules may become brittle in face of unpredictable execution dynamics and can quickly get invalidated. Instead of baseline schedule, another line of work is to consider design of good schedule policies. One such example is the notion of Partial Order Schedules (POS) defined by Policella et al. (2004) which seeks to retain temporal flexibility whenever the problem constraints allow it and can often absorb unexpected deviation from predictive assumptions. They considered robustness measures such as fluidity and flexibility. Generating POS is another example of such flexible approaches: a subset of sequencing decisions are made offline and the remaining decisions are made online by using a dispatching rule (Bidot et al., 2009). Different methods of generating POS were compared in terms of the robustness of the resulting schedules in the work of Rasconi, Cesta, and Policella (2010). In our work, we apply the concept of POS as the execution policy. Given an RCPSP/max instance, mean and variance values of the segregated variables of data perturbations and the level of risk, the objective of our work is to determine POS with a locally minimal robust value.

## 8.3 Scenario-based Optimization in Scheduling

Another line of work that deals with scheduling under uncertainty is based on the use of scenarios (scenario-based optimization). For example, Kouvelis, Daniels, and Vairaktarakis (2000) introduced the concept of robustness into scheduling problems. They considered uncertain processing times and proposed methods to generate a robust schedule based on the maximum absolute deviation between the robust solution against all possible scenarios in a given scenario set. A shortcoming of this kind of approach is that all scenarios are assumed to be known in advance, and that the scenario space is usually exponentially large. Noteworthy of mention are the two notions of solution robustness and quality robustness, where solution robustness (or stability) refers to the insensitivity of actual start times, whereas quality robustness refers to the insensitivity of solution quality (i.e. makespan) to different scenarios (Herroelen & Leus, 2005). Another pioneering scenario-based optimization work is by Mulvey, Vanderbei, and Zenios (1995) which handles the tradeoff between solution robustness (if a solution remains close to the optimal for all scenarios) and model robustness (if a solution remains feasible for most scenarios).

## 8.4 Robust Optimization in Scheduling

A recent development in Operations Research saw the potential of applying the concept of Robust Optimization to deal with uncertainty. Ben-Tal and Nemirovski (2002) and Bertsimas and Sim (2003) proposed robust optimization models where no assumptions of the underlying probability distribution of data are needed. The idea is often to approximate data uncertainty by a tractable (convex) uncertainty set, and optimization is performed on that set. This results in a robust counterpart formulation as a conic (such as second-order cone) optimization problem which can be solved in polynomial time. However, only a few works have been reported in the literature on applying robust optimization to scheduling, due mainly to a high-degree combinational nature of the problem. One such application is the process scheduling problem in chemical engineering, such as the works by Janak, Lin, and Floudas (2007) and Li and Ierapetritou (2008). A notable recent breakthrough in robust optimization on tractable approximation models to solve stochastic optimization problems is found by Chen et al. (2008). This work makes use of linear segregated decision rules that are relevant to solving combinatorial scheduling problems with durational uncertainty and our work exploit this mechanism and incorporate it into local search.

## 9. Conclusion

Given a level of risk $0 < \varepsilon \le 1$ chosen by the planner, we investigated the problem of finding the minimum $(1 - \varepsilon)$-guaranteed makespan (i.e. Robust Makespan) and proposed methods to find a schedule policy (POS) such that when uncertainty is dynamically realized, the execution policy will result in a solution whose value is as good as robust makespan. We first put forward a new decision rule utilized in scheduling to help specify the start times for all activities with respect to execution policy and dynamic realizations of data uncertainty. Based on the decision rule, new fitness function was then derived to evaluate robustness, which was finally integrated into a local search framework to produce the solution with robust makespan. Experimental results illustrate the improved performance of local search with the new fitness evaluation, which provider tighter bounds on robust makespan and better partial order schedules compared to the existing method.

For simplicity we have adopted an upper bound approach where we assume independence among the durational uncertainties. One future work is to treat correlations between durational uncertainties, since a task duration could be correlated with some others in real life. For example, correlations occur when an external event is not peculiar to a single task, but more universal, such as weather conditions, seasonal peaks. In such situations, the durational delays are correlated in the same direction. When this occurs, the decision rules proposed in this paper break down unfortunately, since even if the covariances of pairs of duration variables are given, it is very complex to analytically model the extent to which one duration and any combination (resulting from *SUM* and *MAX* operators) of other durations change together. This in turn complicates the analysis on the variance of the makespan variable, and hence the robust makespan. Extending our work to handle covariances is an interesting future direction.

## Acknowledgments

This paper extends previous research by Lau, Ou, and Xiao (2007) and Fu, Varakantham, and Lau (2010). The authors wish to thank all reviewers for their insightful comments.